\documentclass[preprint,12pt]{elsarticle}

\usepackage{amsmath,amsfonts}
\usepackage{algorithmic}
\usepackage{algorithm}
\usepackage{array}
\usepackage[caption=false,font=normalsize,labelfont=sf,textfont=sf]{subfig}
\usepackage{textcomp}
\usepackage{stfloats}
\usepackage{url}
\usepackage{verbatim}
\usepackage{graphicx}
\usepackage{subcaption}
\usepackage{multirow} 
\usepackage{booktabs} 
\usepackage{pifont} 
\usepackage{amsmath}
\usepackage{color}
\usepackage{xcolor}
\usepackage{subcaption}
\usepackage{caption}
\usepackage{tabularx}
\usepackage{adjustbox}

\newcommand{\cmark}{\ding{51}} 
\newcommand{\xmark}{\ding{55}} 

\usepackage{xcolor}

\journal{Information Fusion}

\begin{document}

\begin{frontmatter}

\title{PairHuman: A High-Fidelity Photographic Dataset for Customized Dual-Person Generation}   

\author[1]{Ting Pan\fnref{dagger}}
\author[2]{Ye Wang\fnref{dagger}}
\author[3]{Peiguang Jing}
\author[2]{Rui Ma}
\author[4]{Zili Yi\corref{cor}}
\ead{yi@nju.edu.cn}
\author[1]{Yu Liu\corref{cor}}
\ead{liuyu@tju.edu.cn}

\affiliation[1]{organization={School of Microelectronics, Tianjin University},
            city={Tianjin},
            postcode={300072}, 
            country={China}}
\affiliation[2]{organization={School of Artificial Intelligence, Jilin University},
            city={Changchun},
            postcode={130012}, 
            country={China}}
\affiliation[3]{organization={School of Electrical and Information Engineering, Tianjin University},
            city={Tianjin},
            postcode={300072}, 
            country={China}}
\affiliation[4]{organization={School of Intelligent Science and Technology, Nanjing University},
            city={Nanjing},
            postcode={215163}, 
            country={China}}

\fntext[dagger]{These authors contributed equally to this work.}
\cortext[cor]{Corresponding author.}

\begin{abstract}
Personalized dual-person portrait customization has considerable potential applications, such as preserving emotional memories and facilitating wedding photography planning. However, the absence of a benchmark dataset hinders the pursuit of high-quality customization in dual-person portrait generation. In this paper, we propose the PairHuman dataset, which is the first large-scale benchmark dataset specifically designed for generating dual-person portraits that meet high photographic standards. The PairHuman dataset contains more than 100K images that capture a variety of scenes, attire, and dual-person interactions, along with rich metadata, including detailed image descriptions, person localization, human keypoints, and attribute tags. We also introduce DHumanDiff, which is a baseline specifically crafted for dual-person portrait generation that features enhanced facial consistency and simultaneously balances in personalized person generation and semantic-driven scene creation. Finally, the experimental results demonstrate that our dataset and method produce highly customized portraits with superior visual quality that are tailored to human preferences. Our dataset is publicly available at https://github.com/annaoooo/PairHuman.
\end{abstract}


\end{frontmatter}

\section{Introduction}

Portraits are vital for capturing individual identities and preserving memories. As one of the most common types of portraits, dual-person portraits stand out for their ability to distinctly convey interpersonal interactions and emotional connections between subjects. Personalized dual-person portrait generation has broad applications. For example, in wedding photography, couples can preview and select sample portraits before the shoot to ensure that their preferences are accurately reflected. In healthcare, personalized portraits of patients with their loved ones serve as valuable tools for reminiscence therapy and emotional support. In psychology and the social sciences, they offer customizable visual stimuli for studying interpersonal dynamics, social behavior, and emotional expression. In human-computer interaction, this technology can support more engaging and personalized content for social platforms and virtual environments. This study aims to advance the generation of personalized dual-person portraits to meet diverse customization needs.

\begin{figure}[!t]
\newcommand{\uwidth}{4.5} 
\newcommand{\uhoriz}{0.07}  
\small
\centering{ %
\begin{minipage}{0.9\textwidth} %
\centering{
\includegraphics[width= \uwidth in]{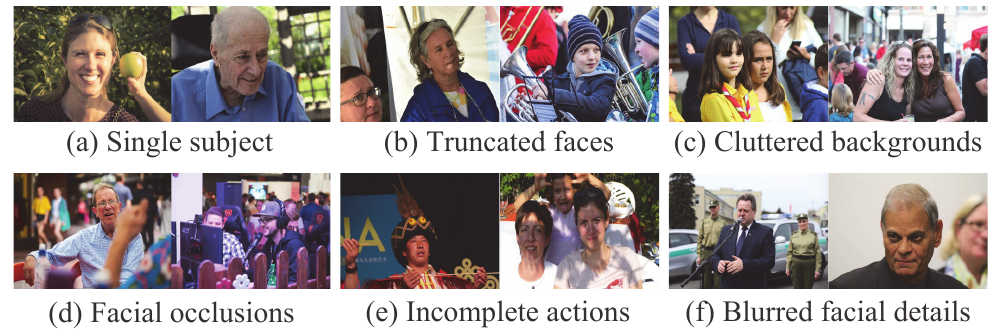}
}
\end{minipage}
}
\vspace{-0.5 em} 
\caption[Instance images from the current multi-person image dataset.]{Examples from the current multi-person image dataset that are unsuitable for high-fidelity dual-person portrait generation\footnotemark.} 
\label{fig:display_error}
\end{figure}
\footnotetext{Image data sourced from the FFHQ-wild dataset.} 

Recent advances in diffusion models \cite{Ho2020ddpm, song2021ddim, Rombach2021LDM, Podell2023SDXL} have greatly enhanced image generation, enabling multi-subject personalization methods such as FastComposer \cite{xiao2024fastcomposer}, FaceDiffuser \cite{wang2024facediffuser}, and MOA \cite{Wang2024MoA}. These methods generate customized images from a shared text prompt and one reference photo per individual. However, generating personalized dual-person portraits presents two key challenges: (i) the lack of diverse, high-quality dual-person portrait datasets, and (ii) the need for more effective high-fidelity personalized generation methods. First, current multi-subject generation image datasets such as FFHQ-wild \cite{Karras2021FFHQ} and ImageNet-1k \cite{Olga2015ImageNet} lack the visual diversity and detailed annotations necessary for customizing dual-person portraits. These datasets contain few dual-person images, are mostly composed of half-body shots, and provide only simple text prompts. Some images also suffer from quality issues, such as truncated faces, cluttered backgrounds, and blurred facial details, as shown in Fig.~\ref{fig:display_error}. These compromise the quality of generated portraits. In addition, generating visual content that is consistent with both guided text prompts and reference images poses another challenge. Specifically, the generation method must accurately align subject identities with intricate text prompts and accurately preserve facial consistency for each subject.

\begin{table*}[!t]
\centering
\setlength{\tabcolsep}{1.5pt} 
\caption{Comparison of different multimodal datasets.}
\vspace{-0.5 em} 
\resizebox{\linewidth}{!}
{
\label{tab:dataset_comparison}
\begin{tabular}{c|cccc|cccccc}
\hline
\multirow{3}{*}{\textbf{Dataset}}  & \multicolumn{4}{c|}{\textbf{Image}} & \multicolumn{6}{c}{\textbf{Annotation}} \\  \cline{2-11}
 & \multicolumn{1}{c}{\textbf{Image}} & \multicolumn{1}{c}{\textbf{Human-}} & \multicolumn{1}{c}{\textbf{Photographic}} & \multicolumn{1}{c|}{\textbf{Two}} & \multicolumn{1}{c}{\textbf{Detailed}} & \multicolumn{1}{c}{\textbf{Attribute}} & \multicolumn{1}{c}{\textbf{Human}} & \multicolumn{1}{c}{\textbf{Human}} & \multicolumn{1}{c}{\textbf{Key-}} & \multicolumn{1}{c}{\textbf{Face}} \\ 
 & \multicolumn{1}{c}{\textbf{Number}}  & \multicolumn{1}{c}{\textbf{Centric}} & \multicolumn{1}{c}{\textbf{Quality}} & \multicolumn{1}{c|}{\textbf{Person}} & \multicolumn{1}{c}{\textbf{Captions}} & \multicolumn{1}{c}{\textbf{Tags}} & \multicolumn{1}{c}{\textbf{BBoxes}} & \multicolumn{1}{c}{\textbf{Masks}}  & \multicolumn{1}{c}{\textbf{points}} & \textbf{BBoxes} \\ 
\hline
Laion400M \cite{Schuhmann2021LAION400M} \& Laion 5B\cite{schuhmann2022laion5b} & 47B/58.5B & \xmark & \xmark & \xmark & \xmark & \xmark & \xmark & \xmark & \xmark & \xmark \\ 
ShareGPT4V \& ShareGPT4V-pt \cite{Chen2023ShareGPT4V} & 100K/1000K & \xmark & \xmark & \xmark & \cmark & \xmark & \xmark & \xmark & \xmark & \xmark \\
RefClef \cite{Kazemzadeh2014ReferItGame} & 19k & \xmark & \xmark & \xmark & \xmark & \xmark & \xmark & \xmark & \xmark & \xmark \\ 
RefCOCO, RefCOCO+, RefCOCOg \cite{Kazemzadeh2014ReferItGame} & 20k-25.8k & \xmark & \xmark & \xmark & \xmark & \cmark & \cmark & \cmark & \cmark & \xmark \\
MS COCO caption \cite{Chen2015COCOcaption} & 328k & \xmark & \xmark & \xmark & \xmark & \cmark & \cmark & \cmark & \cmark & \xmark \\ 
Flickr30K \cite{Plummer2015Flickr30k} & 30k & \xmark & \xmark & \xmark & \xmark & \xmark & \xmark & \xmark & \xmark & \xmark \\
ImageNet-1k \cite{Olga2015ImageNet} & 280k & \xmark & \xmark & \xmark & \xmark & \cmark & \cmark & \cmark & \xmark & \xmark \\
Visual Genome \cite{Krishna2016VisualGC} & 100k & \xmark & \xmark & \xmark & \xmark & \cmark & \cmark & \cmark & \xmark & \xmark \\
Nocaps \cite{Agrawal2019nocaps} & 15k & \xmark & \xmark & \xmark & \xmark & \xmark & \xmark & \xmark & \xmark & \xmark \\ 
FFHQ-wild  \cite{Karras2021FFHQ}\cite{xiao2024fastcomposer}& 70k & \cmark & \xmark & \xmark & \xmark & \cmark & \cmark & \cmark & \xmark & \xmark \\ 
\textbf{PairHuman}  & 100K & \cmark & \cmark & \cmark & \cmark & \cmark & \cmark & \cmark & \cmark & \cmark \\ 
\hline
\end{tabular}
}
\end{table*}

To address existing dataset limitations, we developed PairHuman, which is a high-quality benchmark dataset specifically designed for dual-person portrait generation. The key advantages of PairHuman include the following: (i) Extensive and diverse visual content: PairHuman contains over 100K images and covers four primary topics: couples, weddings, female friends, and parent-child. For each topic, this dataset provides a wide variety of scenes, attire, actions, and compositions. (ii) High photographic fidelity: The dataset provides high-resolution, realistic images that adhere to photographic standards, minimizing issues such as facial truncation. (iii) Standardized dual-person portraits: Every image in PairHuman consistently includes two individuals, ensuring that methods can generate dual-person portraits more accurately. (iv) Rich data annotations: The dataset includes detailed image captions, attribute annotations for individuals and scenes, precise human localization, and human keypoints.\\
\indent To fully leverage the constructed dataset, we developed DHumanDiff. The method is designed to generate high-fidelity dual-person portraits, enabling the reference subjects to be customized across a wide range of scenes, attire, and actions. DHumanDiff employs visual disparity-aware conditioning and combines local and global features to improve facial consistency. Additionally, adapter technology is employed to integrate visual-textual diffusion and achieve more efficient training. During generation, DHumanDiff uses a cascaded inference mechanism to manage scenes and subjects effectively, which allows for adjustable image layouts. The experimental results demonstrate that our dataset and method effectively generate highly customizable dual-person portraits and more accurately preserve individual identities.

The key contributions of this work are as follows:

\begin{enumerate}
\item{We present PairHuman, which is the first large-scale dual-person portrait dataset. It comprises 100K high-fidelity images with diverse visual content and rich annotations. This dataset provides a valuable foundation for advancing personalized dual-person portrait generation.}
\item{We propose DHumanDiff, a method for personalized dual-person portrait generation that enhances facial consistency and supports flexible image layouts. Additionally, the use of an adapter improves training efficiency.}
\item{Our experimental results confirm the necessity of this dataset and demonstrate the effectiveness of the proposed method for generating high-fidelity, diverse dual-person portraits.}
\end{enumerate}

The remainder of this paper is structured as follows: Section II reviews the existing multimodal datasets and personalized generation methods. Section III introduces the PairHuman dataset, outlining its development and presenting statistical analyses. Section IV describes DHumanDiff, which is our method for dual-person portrait generation. Section V presents the experimental results, and Section VI concludes the study.

\section{Related Work}

\subsection{Image Generation Datasets}

In personalized image generation, the quality and annotation of the dataset are critical. A detailed comparison of various multimodal datasets is presented in Table~\ref{tab:dataset_comparison}. Large-scale datasets such as LAION-400M \cite{Schuhmann2021LAION400M} and LAION-5B \cite{schuhmann2022laion5b} offer extensive image‒text pairs, which are automatically curated and filtered using the CLIP model\cite{alec2021clip}. Additionally, ShareGPT4V \cite{Chen2023ShareGPT4V} leverages advanced GPT-4 Vision \cite{2023openai_gpt4v} to generate detailed image captions. Human-centric datasets, including Flickr30K \cite{Plummer2015Flickr30k}, COCO Captions \cite{Chen2015COCOcaption}, RefCOCO \cite{Kazemzadeh2014ReferItGame}, NoCaps \cite{Agrawal2019nocaps}, RefClef \cite{Kazemzadeh2014ReferItGame}, and Visual Genome \cite{Krishna2016VisualGC}, primarily provide brief captions, with some including bounding boxes (BBoxes), masks, and keypoint annotations. However, these datasets focus mainly on general object recognition \cite{zhang2024, Omar1} and scene understanding \cite{ma2025, Omar3}and often lack detailed facial features and human-centric content.
In personalized image generation, the quality and annotation of the dataset are critical. A detailed comparison of various multimodal datasets is presented in Table~\ref{tab:dataset_comparison}. Large-scale datasets such as LAION-400M \cite{Schuhmann2021LAION400M} and LAION-5B \cite{schuhmann2022laion5b} offer extensive image‒text pairs, which are automatically curated and filtered using the CLIP model\cite{alec2021clip}. Additionally, ShareGPT4V \cite{Chen2023ShareGPT4V} leverages advanced GPT-4 Vision \cite{2023openai_gpt4v} to generate detailed image captions. Human-centric datasets, including Flickr30K \cite{Plummer2015Flickr30k}, COCO Captions \cite{Chen2015COCOcaption}, RefCOCO \cite{Kazemzadeh2014ReferItGame}, NoCaps \cite{Agrawal2019nocaps}, RefClef \cite{Kazemzadeh2014ReferItGame}, and Visual Genome \cite{Krishna2016VisualGC}, primarily provide brief captions, with some including bounding boxes (BBoxes), masks, and keypoint annotations. However, these datasets focus mainly on general object recognition \cite{zhang2024, Omar1} and scene understanding \cite{ma2025, Omar3}and often lack detailed facial features and human-centric content.

FFHQ-wild \cite{Karras2021FFHQ} is a high-resolution human image dataset that includes individuals from various ethnicities and age groups, which makes it suitable for human-related computer vision tasks \cite{sun2024}. To address the need for personalized multi-person image generation, Xiao et al. \cite{xiao2024fastcomposer} annotated and aligned image captions with masks within the FFHQ-wild dataset. This work has supported several related studies \cite{xiao2024fastcomposer, wang2024facediffuser, Wang2024MoA}. Despite these contributions, FFHQ-wild lacks diverse dual-person visuals and detailed descriptions, which limits its effectiveness in generating high-fidelity dual-person portraits. Building on previous research, this study develops a high-fidelity dual-person photographic dataset with rich annotations, thereby establishing a foundation for personalized dual-person portrait generation.

\subsection{Personalized Image Generation}
\subsubsection{Single-Subject Personalization} Diffusion models have significantly advanced the generation of realistic facial details and diverse scenes from text inputs\cite{Rombach2021LDM, Ramesh2022DALLE, Saharia2022Imagen}. Building on these advancements, various methods have been developed for subject-specific image generation. Techniques like DreamBooth \cite{Ruiz2023dreambooth} and Textual Inversion \cite{gal2023TextualInversion} personalize a subject's identity by fine-tuning text-to-image models or converting the subject's image feature into a text token. These methods require subject-specific tuning to achieve accurate personalization. Recent methods like IP-Adapter \cite{Ye2023IPAdapter} and InstantID \cite{Wang2024InstantID} incorporate image conditioning and fine-tune text-to-image models during training, thereby enabling tuning-free inference. 

\subsubsection{Multi-Subject Personalization} Extending single-subject personalization methods to multiple subjects introduces challenges, such as identity interference and poor layout. The CustomDiffusion method \cite{Kumari2022Custom} can fine-tune for multiple targets but often struggles to maintain identity consistency within the same category. Layout-based approaches that utilize BBoxes \cite{Inoue2023LayoutDM, Zheng2023LayoutDiffusion} and segmentation maps \cite{Avrahami2022SpaText, Sarukkai2023Collage, Nichol2021GLIDE} assist in generating images with multiple concepts but need a fine-tuned model and additional layout information.

FastComposer \cite{xiao2024fastcomposer} enables tuning-free inference by injecting subject-specific features into argument text tokens. It preserves distinct identities by a cross-attention localization loss. Subsequent methods like FaceDiffuser \cite{wang2024facediffuser} and MoA \cite{Wang2024MoA} advance customization through textual prior retention: FaceDiffuser integrates text-driven and subject-augmented diffusion models at different denoising stages, whereas MoA uses a dual-branch attention and learnable router to balance personalization and prior retention. However, these methods often face challenges in maintaining accurate facial features and demand significant training resources, which underscores the need for more effective strategies for dual-person portrait personalization.

\section{Dataset}

\begin{figure}[!t]
\newcommand{\uwidth}{4.3} 
\newcommand{\uhoriz}{0.07}  
\small
\centering{ %
\begin{minipage}[b]{ 0.9 \textwidth} %
\centering{
\includegraphics[width= \uwidth in]{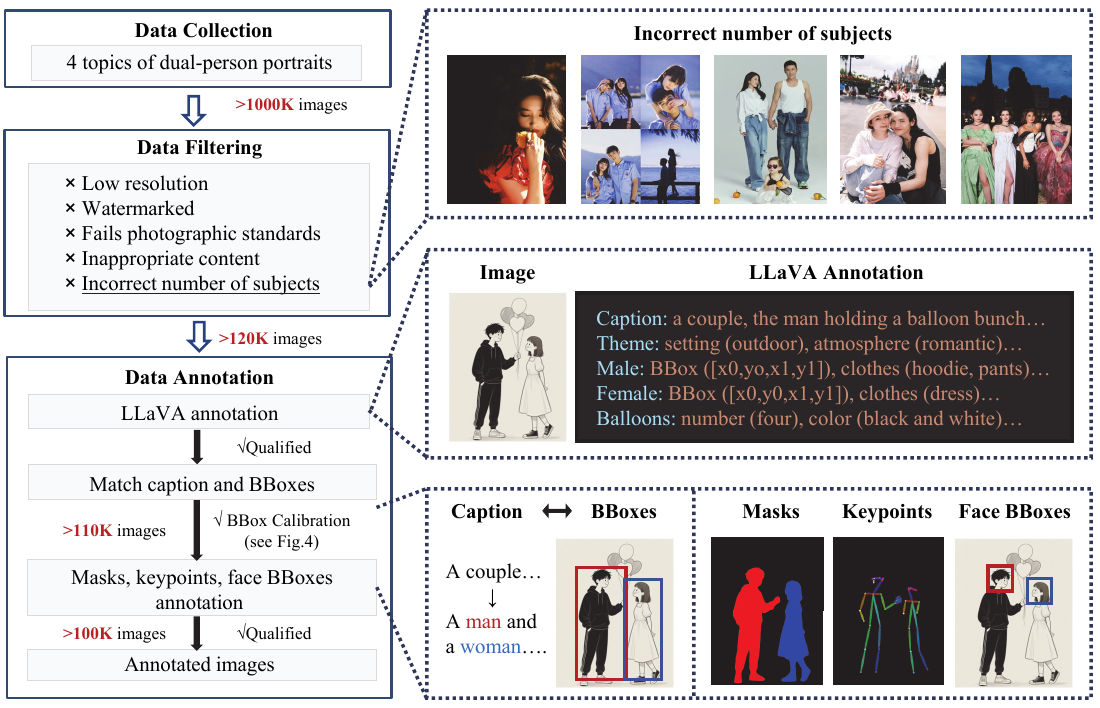}
}
\end{minipage}
}
\vspace{-0.5 em} 
\caption{Illustration of the data collection and annotation process for the PairHuman dataset.}
\vspace{-1em} 
\label{fig:annotation_process}
\end{figure}

\subsection{Data Collection and Filtering}

\begin{figure*}[!t]
\newcommand{\uwidth}{5.4} 
\newcommand{\uhoriz}{0.07}  
\small
\centering{ %
\begin{minipage}[b]{0.98 \textwidth} %
\centering{
\includegraphics[width= \uwidth in]{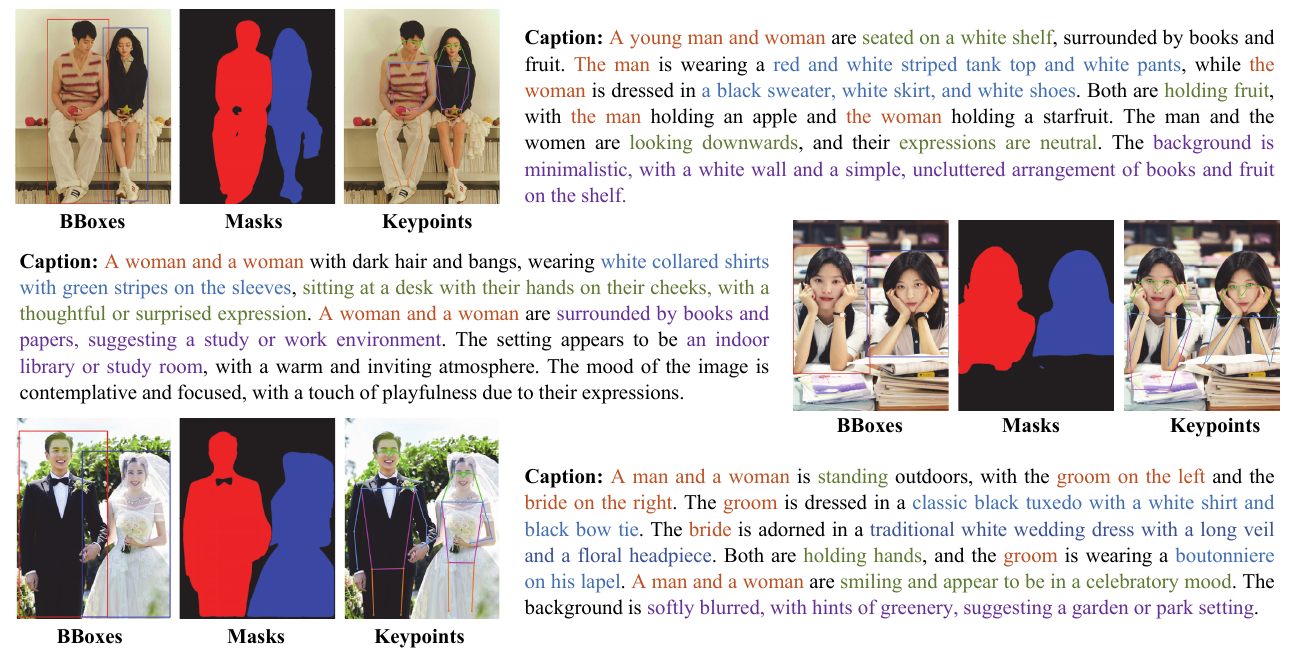}
}
\end{minipage}
}
\vspace{-1.5 em} 
\caption{Examples of PairHuman Dataset Annotations, including human bounding boxes, keypoints, masks, and image captions. Image captions are color-coded for clarity: orange for persons, green for actions, blue for attire, and purple for backgrounds.}
\vspace{-1em}
\label{fig:annotation_display}
\end{figure*}

The data collection process for the PairHuman dataset adheres to a standardized pipeline, as illustrated in Fig.~\ref{fig:annotation_process}. We collected dual-person portraits across four topics: couple, wedding, female friendship, and parent-child. Data collection was guided by more than 50 carefully selected keywords, covering cultural aesthetics, emotional expressions, environmental settings, seasons, and attire. This initial collection yielded over 1000K images. These images were then filtered, annotated (see Section \ref{sec:annotation}), and calibrated (see Section \ref{sec:calibration}), resulting in a dataset of 100K high-quality dual-person portraits with rich annotations. Fig.~\ref{fig:annotation_display} displays examples of the annotated data. Additional examples of the data and detailed keyword contents are provided in \ref{appendix: Data Collection Keywords} and \ref{appendix: Illustrative Examples of the Dataset}.

The data filtering process included: 
(i) Duplicates were removed using hash value analysis \cite{1992hash}. 
(ii) Images that contained specific textual elements, such as ``@'' and ``studio'', were excluded using PaddleOCR \cite{li2022ppocr3}.
(iii) Only images with resolutions greater than 400$\times$400 pixels were retained, with over 90\% exceeding 1024$\times$1024 pixels.
(iv) Dual-person portraits were screened using Fast R-CNN \cite{Girshick2015fastrcnn}, identifying approximately 120K images that contained two people. These were further verified by the annotations of human BBoxes, masks, and keypoints, resulting in a final set of 100K dual-person portraits.
(v) Finally, a manual inspection was conducted to remove images that did not meet photographic aesthetic standards or contained inappropriate content.

\subsection{Data Annotation}
\label{sec:annotation}
Accurate semantic scene understanding \cite{wang2025, Omar2} and human-centered annotation are crucial for customized portrait generation. To achieve this objective, we conducted extensive multimodal data annotations, which include detailed image captions, attribute tags, human BBoxes, masks, keypoints, and facial BBoxes. The annotation methods are described below.

Image captions, photo themes, and BBoxes with attributes for individuals and objects were generated using the LLaVA-next method \cite{liu2024llavanext, li2024llavanextstrong}. To address response length constraints inherent to the LLaVA-next model’s processing capabilities, we designed three specialized prompts, as illustrated in Fig.~\ref{fig:annotation_process}. Notably, the second prompt utilized LLaVA’s grounding capabilities to generate initial BBoxes and the corresponding attribute annotations for each individual.Further details on the annotation prompts are provided in the \ref{appendix: Prompts of LLaVA annotation}.

To match each individual's identity accurately with their location in the image, we converted generic descriptors such as ``couple'' into specific terms such as ``a man and a woman'' inspired by the FFHQ-wild annotation method \cite{xiao2024fastcomposer}. Given that human BBoxes and image captions were annotated separately, we employed a greedy matching algorithm that uses gender, relationship, and positional cues to align BBox annotations with the corresponding descriptions in the captions.

The following annotation methods for masks, keypoints, and facial BBoxes were used: Mask annotations were generated using OneFormer \cite{Jain2023oneformer}, with Mask2Former \cite{Cheng2022mask2former} as an alternative. Keypoints were annotated using ViTPose \cite{Xu2024vitpose} on the basis of the calibrated BBoxes. Facial BBoxes were centered at the midpoint between the eyes, with dimensions set to four times the eye-ear distance to fully encompass the facial features.

\begin{figure}[!t]
\newcommand{\uwidth}{5} 
\newcommand{\uhoriz}{0.07}  
\centering
\scriptsize
\begin{minipage}[t]{0.98\textwidth}
\begin{algorithm}[H]
\renewcommand{\thealgorithm}{}
\caption{Human Bounding Boxes Calibration}\label{alg:alg1}
\footnotesize
\begin{algorithmic}
\STATE \textbf{Input} LLaVA grounding result, six object detection results
\STATE \textbf{if} IoU $(\text{BBox}_1, \text{BBox}_2) < 0.8$ in LLaVA result \textbf{then}
\STATE \hspace{0.5cm} \textbf{if} count(person BBoxes) = 2 in more than two detections \textbf{then}
\STATE \hspace{1cm} Match BBox pairs by Eq.1
\STATE \hspace{1cm} Identify the best-matching detection results by Eq.2
\STATE \hspace{1cm} Determine ultimate BBoxes pair by Eq.3
\STATE \hspace{0.5cm} \textbf{end if}
\STATE \textbf{end if}
\STATE \textbf{Output} calibrated result
\end{algorithmic}
\end{algorithm}
\end{minipage}
\vspace{0.2em}
\hfill
\\
\begin{minipage}[b]{0.98\textwidth}
\centering
\includegraphics[width=\uwidth in]{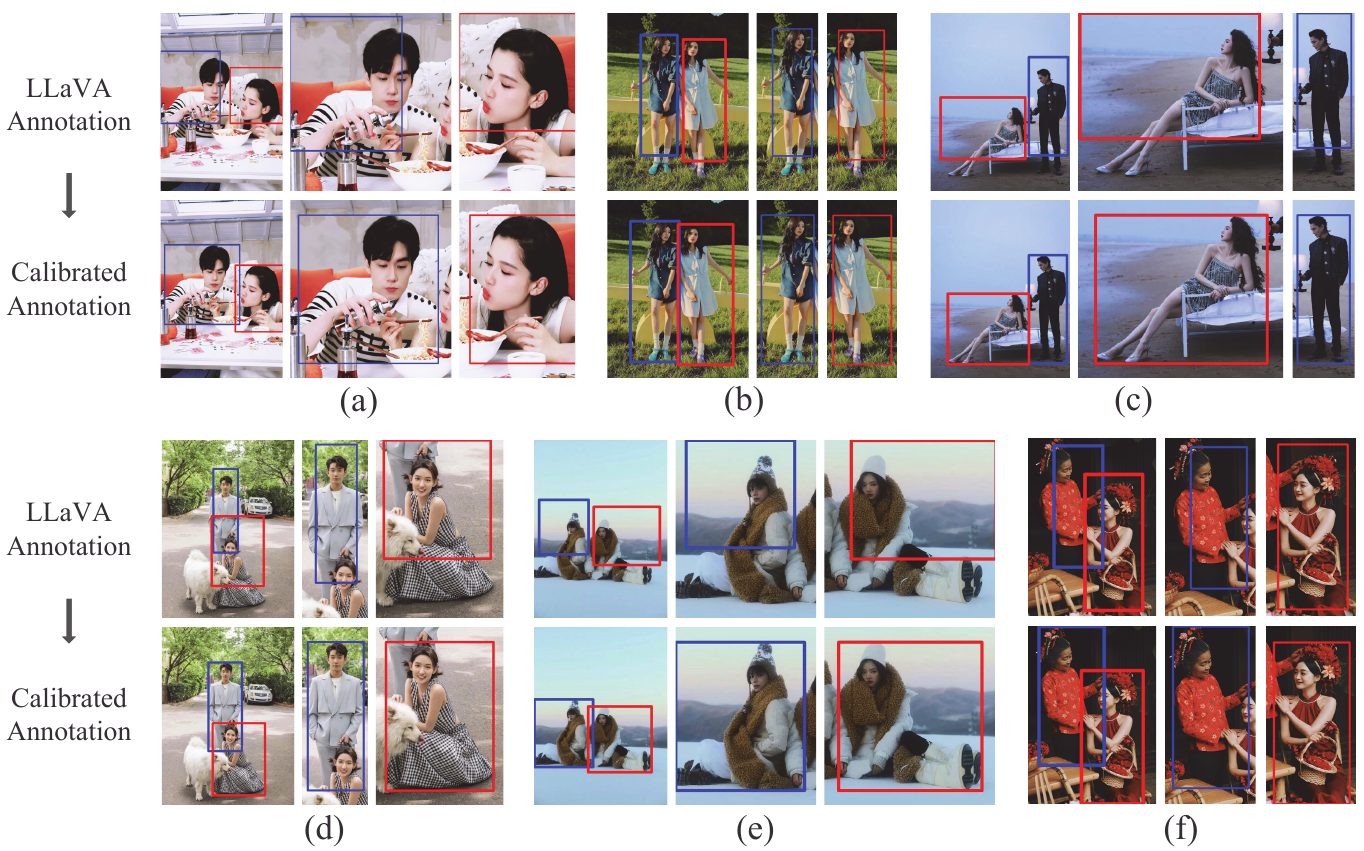}
\end{minipage}
\vspace{-0.8em}
\caption{Calibration of LLaVA's human localization bounding boxes using six object detection methods.}
\vspace{-1em}
\label{fig:BBox_Calibration}
\end{figure}

\subsection{Annotation Validation} 
\label{sec:calibration}
\subsubsection{Validation of Annotations} After completing the LLaVA annotation process, we conducted manual verification for both image captions and attribute descriptions. Next, we set the intersection over union (IoU) threshold below 0.8 to ensure accurate localization, as our experiments confirmed that this threshold effectively prevents duplicate detection of the same individual. Any annotation errors identified during this process were reannotated. Additionally, the precision of the image captions, attribute descriptions, BBoxes, masks, keypoints, and facial BBoxes was rigorously validated through random manual quality checks.

\subsubsection{Bounding Box Calibration}
To address inaccuracies in LLaVA’s location results, six advanced object detection methods were employed for calibration: YOLOv8 \cite{Reis2023yolov8}, YOLOv9 \cite{Wang2024yolov9}, YOLOX-x \cite{ge2021yolox}, PP-Human \cite{ma2024pphuman}, RT-DETR \cite{zhao2023rtdetr}, and MaskDINO \cite{li2022maskdino}. As illustrated in the algorithm in Fig.~\ref{fig:BBox_Calibration}, the calibration process involved the following steps:

First, LLaVA detections with an IoU score below 0.8 were retained. Among the six object detection results, only those with two person-class BBoxes were selected. To align the detection results across different algorithms, the mean squared error (MSE) between each pair of BBoxes was calculated. The pair with the lowest MSE was assigned to one person, and the remaining pair was automatically matched, as defined by:
\begin{equation} \small
\underset{i_1, i_2 \in \{1,2\}}{\text{arg min}} \left[ \text{MSE}(\mathbf{BBox}^{i_1}_a, \mathbf{BBox}^{i_2}_b) \right]  \label{eq1}
\end{equation}
where $a$ and $b$ represent two different algorithms, and $i_1$ and $i_2$ denote BBox indices.

Next, the two most accurate detection algorithms, ${opt_1}$ and ${opt_2}$, were identified by minimizing the average MSE between BBoxes for the same person:
\begin{equation} \small
\underset{u_1, u_2 \in \{1,\ldots,6\}, {u_1} \neq {u_2}}{\text{arg min}} \left[ \frac{1}{2} \sum_{i=1}^{2} \text{MSE}(\mathbf{BBox}^{i}_{u_1}, \mathbf{BBox}^{i}_{u_2}) \right] \label{eq2}
\end{equation}
where ${u_1}$ and ${u_2}$ denote different object detection algorithms.

Finally, the calibrated BBoxes were determined by minimizing the MSE between the detections of LLaVA and those of the two optimal object detection algorithms, expressed as:
\begin{equation} \small
\underset{v \in \{{opt}_1, {opt}_2\}}{\text{arg min}} \; \frac{1}{2} \sum_{i=1}^{2} \text{MSE}(\mathbf{BBox}^i_{\text{LLaVA}}, \mathbf{BBox}^i_{v}) \label{eq3}
\end{equation}

During calibration, cases that exhibited significant mismatches in the detection results (IoU among the six object detection algorithms $<$ 0.6, as defined in Eq.\ref{eq2}) or inconsistent calibration (IoU between the object detection algorithms and LLaVA $<$ 0.6, as defined in Eq.\ref{eq3}) were manually reviewed and excluded to ensure annotation reliability.

Fig.~\ref{fig:BBox_Calibration} presents a visual comparison of the localization results before and after calibration. Fig.~\ref{fig:BBox_Calibration}(a) shows the initial BBoxes from LLaVA, which roughly identify the described individuals but often lack precise localization. In contrast, Fig.~\ref{fig:BBox_Calibration}(b) displays the calibrated BBoxes, which are tightly aligned with the actual positions, demonstrating the effectiveness of the calibration method.

\begin{figure*}[!t]
\newcommand{\uwidth}{5.5} 
\small
\centering{ %
\includegraphics[width=\uwidth in]{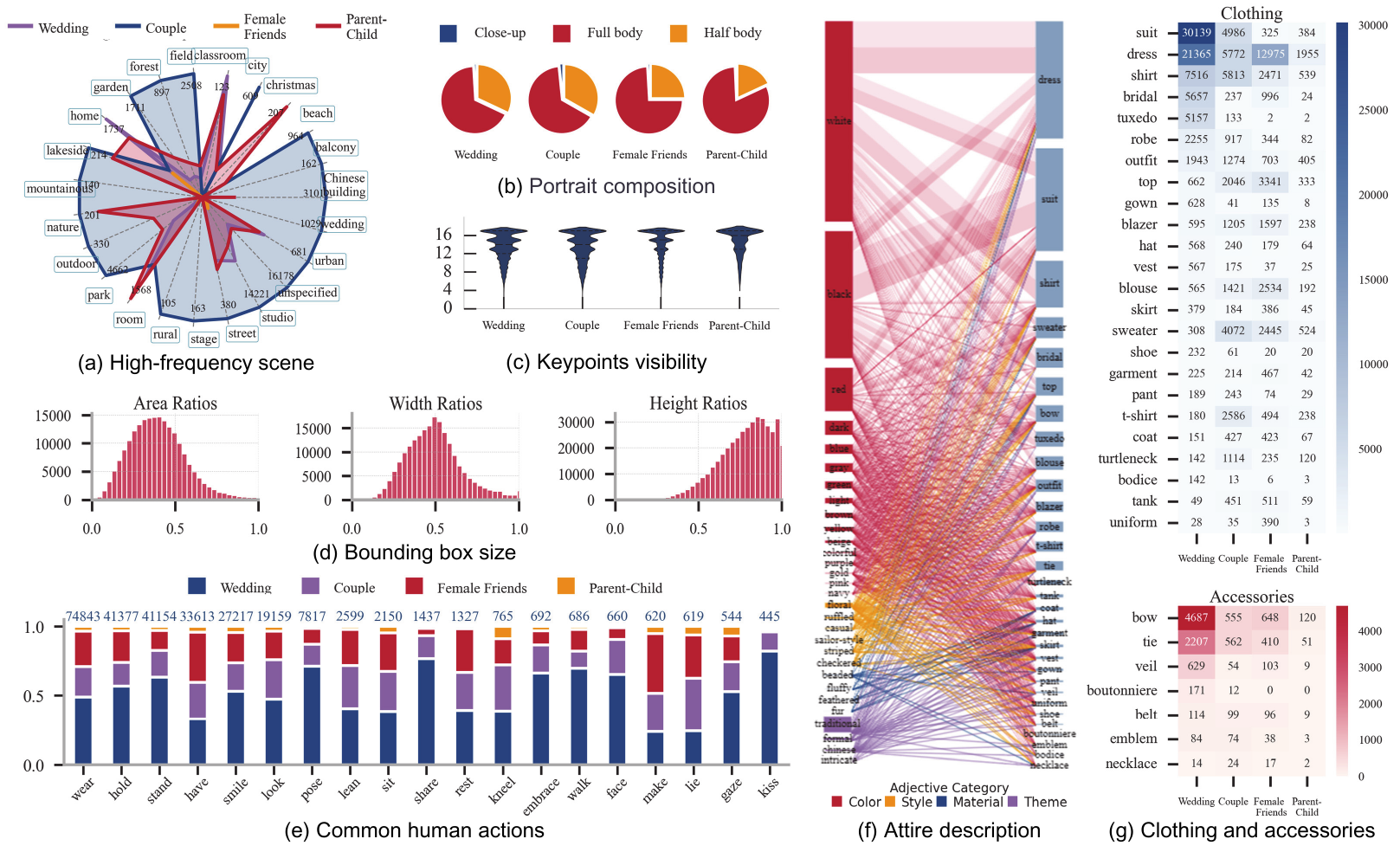}
\vspace{-1.5 em} 
\caption{Multivariate statistical analysis of the PairHuman dataset: (a) Distribution of high-frequency shooting scenes; (b) Distribution of full-body, half-body, and close-up portraits; (c) Visibility distribution of human keypoints; (d) Ratio distribution of human bounding box area, height, and width relative to the entire image; (e) Distribution of common human actions; (f) Correspondence between attire items and descriptive adjectives; (g) Frequency distribution of clothing and accessories.}
\label{fig:data_distribution} 
}
\vspace{-1em} 
\end{figure*}

\subsection{Dataset Partitioning and Statistical Analysis}
\label{sec:statistical_analysis}

\subsubsection{Dataset Division} 

The PairHuman dataset consists of four topics: 51.9\% wedding, 21.9\% couples, 22.5\% female friends, and 3.8\% parent‒child portraits. The smaller proportion of parent‒child portraits is due to restrictions on dual-person portraits. We divided the data into a training set and a test set at a 9:1 ratio with stratified sampling to ensure balanced distributions across the four portrait topics, compositions, and scenes. This division is aligned with standard photography practices for capturing personal relationships, varied compositions, and diverse environments.

\subsubsection{Demographic Statistics}

The PairHuman dataset comprises 201,922 Asian portrait subjects, with a gender distribution of 37.69\% men and 62.31\% women. The age distribution is predominantly composed of young adults, with an average age of 28.3 years (standard deviation: 5.3 years). Most subjects fall within the 20-29 age range (54.98\%), followed by 43.22\% in the 30-39 age range. Other age groups are substantially underrepresented: 0.44\% aged 10-19, 1.33\% aged 40-49, and only 0.03\% aged 50-59. This bias stems from the dataset’s original curation focus on artistic portrait photography, a domain characterized by a predominance of young subjects.

\subsubsection{Shooting Scene}
Fig.~\ref{fig:data_distribution}(a) illustrates the distribution of high-frequency shooting scenes in the PairHuman dataset. The studio settings are the most numerous, which indicates the prevalence of indoor photography. Scenes labeled as ``unspecified,'' which are typically characterized by simple or blurred backgrounds, are also common. The dataset covers a wide range of environments, including natural and residential settings, as well as specialized scenes such as Chinese festivities, Christmas, and weddings.

\subsubsection{Portrait Composition and Keypoint Visibility}
Images were classified into full-body, half-body, and close-up portraits on the basis of the visibility of the shoulder and knee keypoints. Fig.~\ref{fig:data_distribution}(b) shows the distribution of these categories, with full-body images being the most common, followed by half-body portraits, and close-ups being the least frequent. To address this imbalance, we introduced random cropping during data preprocessing, which enhances composition diversity.
Fig.~\ref{fig:data_distribution}(c) presents the keypoint visibility across different portrait topics. Couples and wedding portraits show high consistency due to uniform posing styles. In contrast, parent‒child portraits present greater keypoint visibility, as they often capture the full bodies of children.

\subsubsection{Analysis of Human Bounding Boxes}
Fig.~\ref{fig:data_distribution}(d) analyzes the sizes of human BBoxes and shows that most BBoxes cover 20\% to 50\% of the image, with fewer exceeding half of the image size. The width ratios range primarily from 40\% to 60\%, which is typical for dual portraits, whereas the height ratios peak around 0.8, emphasizing a focus on individual portraits.

\subsubsection{Human Action Analysis}
Fig.~\ref{fig:data_distribution}(e) presents the verbs that describe activities and states within the dataset. The verb ``wear'' is the most frequent, which underscores the dataset's emphasis on attire. Verbs such as ``hold,'' ``stand,'' ``sit,'' and ``lean'' correspond to common portrait poses, whereas ``smile'' and ``look'' capture facial expressions, and ``embrace'' and ``kiss'' indicate human interactions, collectively showcasing the dataset's diverse range of human activities and interactions.

\subsubsection{Attire Attribute Analysis}
Fig.~\ref{fig:data_distribution}(f) shows the relationships between attire items and descriptive adjectives categorized by color, material, style, and theme, with connecting lines indicating the frequency of specific adjective-noun pairings. More detailed statistics are provided in Table~E.5 in the Appendix. Fig.~\ref{fig:data_distribution}(g) displays the distributions of clothing and accessories across portrait topics. Together, these figures illustrate the dataset's diversity in attire types and attribute details.

\section{The Proposed DHumanDiff Method}
\subsection{Problem Formulation}

\begin{figure*}[!t]
\newcommand{\uwidth}{5.5} 
\newcommand{\uhoriz}{0.07}  
\small
\centering{ %
\includegraphics[width= \uwidth in]{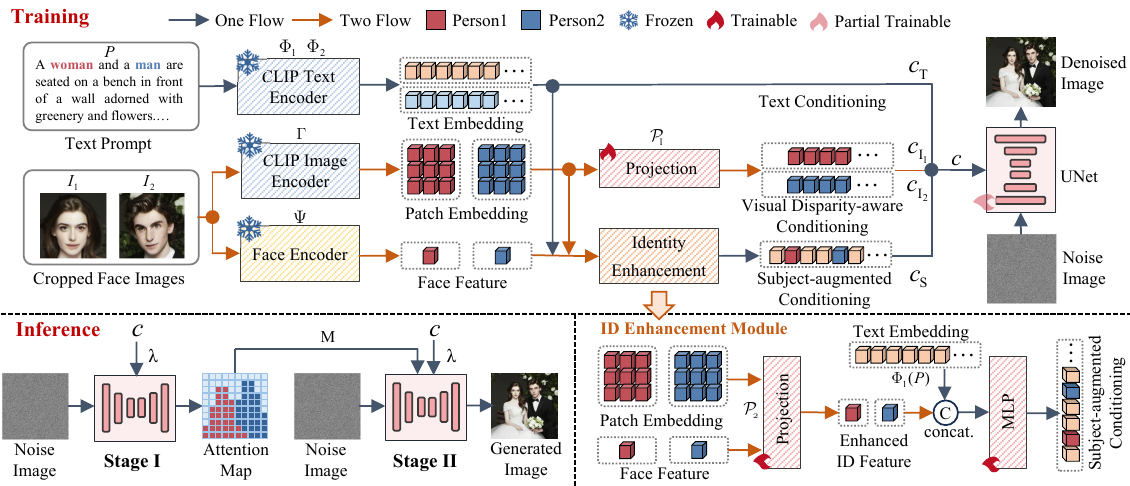}
}
\vspace{-1.5 em} 

\caption{Proposed Framework of DHumanDiff. DHumanDiff generates dual-person portraits from a text prompt and two facial images. It employs subject-augmented conditioning $c_{S}$ with an identity enhancement module and visual disparity-aware conditioning $c_{I_1}$ and $c_{I_2}$ to improve facial consistency (Section 4.2). All conditioning $\mathbf{c} \!=\! \{\mathbf{c}_T, \mathbf{c}_{I_1}, \mathbf{c}_{I_2}, \mathbf{c}_S\}$ is integrated through adapter technology within the diffusion process, enabling effective customization and efficient training (Section 4.3). Additionally, a cascaded inference mechanism balances the subjects generation and the semantic scene creation (Section 4.4).}
\vspace{-1em} 
\label{fig:algorithm_framework}
\end{figure*}

The overall framework of the proposed method is illustrated in Fig.~\ref{fig:algorithm_framework}. Specifically, DHumanDiff includes two CLIP text encoders $\Phi_1$ and $\Phi_2$ for handling the text prompt $P$ and a CLIP image encoder $\Psi$ and a facial encoder $\Gamma$ for handling reference facial images $I_1$ and $I_2$. It further generates four types of conditioning: text conditioning $\mathbf{c}_T$, image conditioning $\mathbf{c}_{I_1}$, $\mathbf{c}_{I_2}$, and subject-augmented conditioning $\mathbf{c}_S$. These conditionings collectively guide the diffusion model to generate customized dual-person portraits.

During training, the grounded dual-person images $\mathbf{x}$ are compressed into a latent representation $\mathbf{z}_0 = \mathcal{E} (\mathbf{x})$ using a variational autoencoder (VAE) encoder $\mathcal{E}$. In the forward diffusion process, Gaussian noise $\boldsymbol{\epsilon}$ is gradually added to the latent code $\mathbf{z}_0$ over timesteps $t \in {1, \dots, T}$, transforming it into a noisy latent variable $\mathbf{z}_T$ that approximates a standard Gaussian distribution:
\begin{equation} \small
    \mathbf{z}_t = \sqrt{\bar{\alpha}_t} \mathbf{z}_0 + \sqrt{1 - \bar{\alpha}_t} \boldsymbol{\epsilon}
\end{equation}
where $\bar{\alpha}_t$ is the cumulative product of $\alpha_t$ up to timestep $t$, and $\alpha_t$ is the noise schedule coefficient at timestep $t$. A UNet-based denoiser $\boldsymbol{\epsilon}_\theta(\mathbf{z}_t, t, c)$ is trained to predict the added noise from the noisy latent $\mathbf{z}_t$ at each timestep $t$, conditioned on $\mathbf{c} = \{\mathbf{c}_T, \mathbf{c}_{I_1}, \mathbf{c}_{I_2}, \mathbf{c}_S\}$.

During inference, we adopt classifier-free guidance~\cite{ho2021classifierfree}, the guided noise prediction is computed as:
\begin{equation} \small
\hat{\boldsymbol{\epsilon}_t} = \boldsymbol{\epsilon}_\theta(\mathbf{z}_t, t, \emptyset) + w \cdot \left( \boldsymbol{\epsilon}_\theta(\mathbf{z}_t, t, c) - \boldsymbol{\epsilon}_\theta(\mathbf{z}_t, t, \emptyset) \right)
\end{equation}
where $w$ is the guidance scale, $\emptyset$ denotes without any conditioning inputs, and $\boldsymbol{\epsilon}_\theta(\mathbf{z}_t, t, \emptyset)$ and $\boldsymbol{\epsilon}_\theta(\mathbf{z}_t, t, \mathbf{c})$ correspond to the unconditional and conditional noise predictions. The latent variable is then updated as:
\begin{equation} \small
\mathbf{z}_{t-1} = \frac{1}{\sqrt{\alpha_t}} \left( \mathbf{z}_t - \frac{1 - \alpha_t}{\sqrt{1 - \bar{\alpha}_t}} \hat{\boldsymbol{\epsilon}_t} \right) + \sigma_t \boldsymbol{\epsilon}
\end{equation}
where $\sigma_t$ is the standard deviation of the noise added at timestep $t$, determined by the noise schedule $\sigma_t$.

This denoising process is performed iteratively from $t=T$ to $t=0$, progressively refining the noisy latent variable $\mathbf{z}_T$ towards a clean latent representation $\mathbf{z}_0$. Finally, the VAE decoder $\mathcal{D}$ maps the recovered latent code $\mathbf{z}_0$ back to the image space to produce the final output image $\hat{\mathbf{x}} = \mathcal{D}(\mathbf{z}_0)$.

\begin{figure*}[!t]
\newcommand{\uwidth}{5} 
\newcommand{\uhoriz}{0.07}  
\small
\centering{ %
\includegraphics[width= \uwidth in]{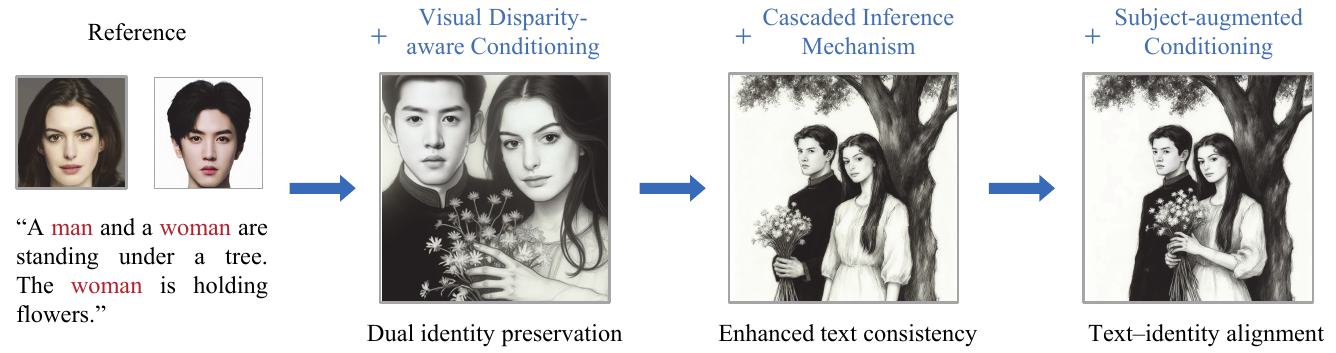}
}
\vspace{-0.5 em} 

\caption{Roles of the core modules in the DHumanDiff framework.}
\vspace{-1em} 
\label{fig:algorithm_conditioning_effect}
\end{figure*}

Fig.~\ref{fig:algorithm_conditioning_effect} illustrates the effects of the core modules in DHumanDiff: (i) visual disparity-aware conditioning introduces image-based conditioning while suppressing inter-subject feature mixing to preserve both identities; (ii) cascaded inference modulates the guidance strength of image-based conditioning across denoising stages, reducing undue emphasis on facial regions and improving semantic consistency; and (iii) subject-augmented conditioning binds the target subject to the prompt, mitigating text–identity mismatches and enabling accurate personalization. Subsequent subsections detail the implementation of these modules and the integration of the various conditioning strategies.

\subsection{Identity Preservation}

\subsubsection{Visual Disparity-Aware Conditioning} DHumanDiff uses a CLIP image encoder $\Psi$ to extract dense patch-level features from each reference image. Due to the substantial volume of these features, they are compressed into token matrices using a transformer-based projector $\mathcal{P}_1$, which results in visual disparity-aware conditioning $\mathbf{c}_{I_1}$ and $\mathbf{c}_{I_2}$:
\begin{equation} \small
\begin{cases}
  \mathbf{c}_{I_1} = \mathcal{P}_1(\Psi(\mathbf{I}_1)) \\
  \mathbf{c}_{I_2} = \mathcal{P}_1(\Psi(\mathbf{I}_2))
\end{cases}
\end{equation}

Specifically, $\mathbf{c}_{I_i}^0$ (for $i \in \{1, 2\}$) are initialized as learnable image condition tokens, each composed of 16 latent tokens with dimensionality matching $\mathbf{c}_T$. Raw CLIP features $\Psi(\mathbf{I}_i)$ are first projected via a learned linear mapping to obtain the projected CLIP features $\mathbf{f}_i^{{clip}_1} = \Psi(\mathbf{I}_i) \mathbf{W}_p^{\mathcal{P}_1}$, where $\mathbf{W}_p^{\mathcal{P}_1}$ is a projection matrix. The cross-attention mechanism in $\mathcal{P}_1$ is represented as:
\begin{equation} \small
\mathbf{A}_{\mathcal{P}_1}^i = \text{Softmax} \left( \frac{\mathbf{Q}_{\mathcal{P}_1}^i (\mathbf{K}_{\mathcal{P}_1}^i)^\top}{\sqrt{d}} \right) \mathbf{V}_{\mathcal{P}_1}^i
\end{equation}
where the query, key, and value matrices are given by $\mathbf{Q}_{\mathcal{P}_1}^i = \mathbf{c}_{I_i}^0 \, \mathbf{W}_{q_{\mathcal{P}_1}},  \mathbf{K}_{\mathcal{P}_1}^i = \left[ \mathbf{f}_i^{{clip}_1} \, || \, \mathbf{c}_{I_i}^0\right] \, \mathbf{W}_{k_{\mathcal{P}_1}}, \mathbf{V}_{\mathcal{P}_1}^i = \left[  \mathbf{f}_i^{{clip}_1} \, || \, \mathbf{c}_{I_i}^0 \right] \, \mathbf{W}_{v_{\mathcal{P}_1}}$. Here, $\mathbf{W}_{q_{\mathcal{P}_1}}$, $ \mathbf{W}_{k_{\mathcal{P}_1}}$, and $\mathbf{W}_{v_{\mathcal{P}_1}}$ are weight matrices used to obtain the query, key, and value representations, respectively. The symbol $||$ denotes concatenation. The attention output is processed through a two-layer feed-forward network with a residual connection, followed by an output projection:
\begin{equation} \small
\mathbf{c}_{I_i} = \left( \text{GELU}\left( \mathbf{A}_{\mathcal{P}_1}^i \mathbf{W}_{1}^{\mathcal{P}_1} \right) \mathbf{W}_{2}^{\mathcal{P}_1} + \mathbf{A}_{\mathcal{P}_1}^i \right)  \mathbf{W}_{o}^{\mathcal{P}_1}
\end{equation}
where $\mathbf{W}_{1}^{\mathcal{P}_1}$ and $\mathbf{W}_{2}^{\mathcal{P}_1}$ are linear layers of the feed-forward network, $\mathbf{W}_{o}^{\mathcal{P}_1}$ is the output projection matrix, and the GELU is activation function applied to introduce non-linearity.

To mitigate identity confusion between reference faces, we extend the original U-Net architecture by integrating two additional cross-attention layers that incorporate $\mathbf{c}_{I_1}$ and $\mathbf{c}_{I_2}$. This integration introduces two separate image-oriented attention mechanisms, described as:
\begin{equation} \small
\begin{cases}
\mathbf{A}_{I_1} \! & = \text{Attention}(\mathbf{Q}, \mathbf{K}_{I_1}, \mathbf{V}_{I_1}) = \text{Softmax}\left(\frac{\mathbf{Q}(\mathbf{K}_{I_1})^T}{\sqrt{d}}\right)\mathbf{V}_{I_1} \\
\mathbf{A}_{I_2} \! & =  \text{Attention}(\mathbf{Q}, \mathbf{K}_{I_2}, \mathbf{V}_{I_2}) = \text{Softmax}\left(\frac{\mathbf{Q}(\mathbf{K}_{I_2})^T}{\sqrt{d}}\right)\mathbf{V}_{I_2}
\end{cases}
\end{equation}
where $\mathbf{A}_{I_1}$ and $\mathbf{A}_{I_2}$ are attention outputs, the query matrix $\mathbf{Q} = \mathbf{z}_t \mathbf{W}_q$ is derived from the latent representation $\mathbf{z}_t$ and interacts with key matrices $\mathbf{K}_{I_1} = \mathbf{c}_{I_1} \mathbf{W}_{k_{I_1}}$ and $\mathbf{K}_{I_2} = \mathbf{c}_{I_2} \mathbf{W}_{k_{I_2}}$ and value matrices $\mathbf{V}_{I_1} = \mathbf{c}_{I_1} \mathbf{W}_{v_{I_1}}$ and $\mathbf{V}_{I_2} = \mathbf{c}_{I_2} \mathbf{W}_{v_{I_2}}$. Here, $\mathbf{W}_q$, $\mathbf{W}_{k_{I_1}}$, $\mathbf{W}_{k_{I_2}}$, $\mathbf{W}_{v_{I_1}}$, and $\mathbf{W}_{v_{I_2}}$ are weight matrices.
 
This additional cross-attention layer also decouples cross-attention between image-driven and text-driven diffusion to avoid joint learning under different conditioning and generate more accurate and refined facial reconstructions.

\subsubsection{Subject-augmented conditioning} The text prompt is defined as $P = \{w_1, \ldots, w_n, \ldots, w_N\}$, where $ n = 1, \ldots, N$ is the index of each word. The indices $m_1$ and $m_2$ are assigned to specific words such as ``man'' and ``woman,'' which correspond to the reference facial images $\mathbf{I}_1$ and $\mathbf{I}_2$, respectively.

Unlike previous methods that rely primarily on global features, \textit{i.e.}, CLS token features extracted from CLIP image encoders \cite{xiao2024fastcomposer} \cite{wang2024facediffuser}, the DHumanDiff method integrates both local and global facial features. The InsightFace encoder $\Gamma$ specializes in capturing local facial details, whereas the CLIP image encoder $\Psi$ focuses on extracting global features. These features are then aligned using a transformer-based projector $\mathcal{P}_2$:
\begin{equation} \small
\begin{cases}
\mathbf{f}_{\text{align}_{m_1}} = \mathcal{P}_2\left( \Gamma(\mathbf{I}_1), \Psi(\mathbf{I}_1) \right) \\
\mathbf{f}_{\text{align}_{m_2}} = \mathcal{P}_2\left( \Gamma(\mathbf{I}_2), \Psi(\mathbf{I}_2) \right)
\end{cases}
\end{equation}
where $\mathbf{f}_{\text{align}_{m_1}}$ and $\mathbf{f}_{\text{align}_{m_2}}$ denote the aligned facial features. Specifically, the identity feature $\mathbf{f}_i^{id}$ is derived from $\Gamma(\mathbf{I}_i)$ via a two-layer projection.
\begin{equation} \small
\mathbf{f}_i^{id} = \text{GELU}(\Gamma(\mathbf{I}_i) \cdot \mathbf{W}_1^{\mathcal{P}_2} ) \cdot \mathbf{W}_2^{\mathcal{P}_2}
\end{equation} 
where $\mathbf{f}_i^{id}$ is projected identity features, and $\mathbf{W}_1^{\mathcal{P}_2}$ and $\mathbf{W}_2^{\mathcal{P}_2}$ are feed-forward layer weights. The $ \Psi(\mathbf{I}_i)$ are linearly mapped to obtain projected CLIP features $ \mathbf{f}_i^{{clip}_2} = \Psi(\mathbf{I}_i) \mathbf{W}_p^{\mathcal{P}_2}$, where $\mathbf{W}_p^{\mathcal{P}_2}$ denotes a projection matrix. The cross-attention operation in $\mathcal{P}_2$ is defined as:
\begin{equation} \small
\mathbf{A}_{\mathcal{P}_2}^i = \text{Softmax} \left( \frac{\mathbf{Q}_{\mathcal{P}_2}^i (\mathbf{K}_{\mathcal{P}_2}^i)^\top}{\sqrt{d}} \right) \mathbf{V}_{\mathcal{P}_2}^i
\end{equation}
where the query, key, and value matrices are defined as $\mathbf{Q}_{\mathcal{P}_2}^i = \mathbf{f}_i^{id} \mathbf{W}_{q_{\mathcal{P}_2}}$, $\mathbf{K}_{\mathcal{P}_2}^i = [\mathbf{f}_i^{id} \| \mathbf{f}_i^{{clip}_2}]
\mathbf{W}_{k_{\mathcal{P}_2}}$, and $\mathbf{V}_{\mathcal{P}_2}^i = [\mathbf{f}_i^{id} \| \mathbf{f}_i^{{clip}_2}] \mathbf{W}_{v_{\mathcal{P}_2}}$. $\mathbf{W}_{q_{\mathcal{P}_2}}$, $ \mathbf{W}_{k_{\mathcal{P}_2}}$, and $\mathbf{W}_{v_{\mathcal{P}_2}}$ are weight matrices. Finally, the output is obtained by:
\begin{equation} \small
\mathbf{f}_{\text{align}_n} = \left( \text{GELU} \left( \mathbf{A}_{\mathcal{P}_2}^i \mathbf{W}_{3}^{\mathcal{P}_2} \right)\mathbf{W}_{4}^{\mathcal{P}_2} + \mathbf{A}_{\mathcal{P}_2}^i \right) \mathbf{W}_o^{\mathcal{P}_2}
\end{equation}
where $n \in \{m_1, m_2\}$, $\mathbf{W}_{3}^{\mathcal{P}_2}$ and $\mathbf{W}_{4}^{\mathcal{P}_2}$ are feed-forward layer weights, and $\mathbf{W}_o^{\mathcal{P}_2}$ is the output projection matrix.

The aligned facial features are integrated into the text embedding at the specified indices via a two-layer multilayer perceptron (MLP) to produce subject-augmented conditioning $\mathbf{c}_S$:
\begin{equation} \small
\mathbf{c}_S^n = 
\begin{cases} 
\Phi_1(w_n) & \text{if } n \notin \{m_1, m_2\} \\
\mathbf{MLP}(\Phi_1(w_n) \, || \, \mathbf{f}_{\text{align}_n}) & \text{if } n \in \{m_1, m_2\}
\end{cases}
\end{equation}
where $\Phi_1$ represents the CLIP text encoder.

Subject-augmented conditioning ensures precise alignment between each subject's unique facial features and their corresponding textual descriptions. By integrating both local and global features, this approach enhances facial consistency. Thus, it ensures that the model's outputs are both high-fidelity and precisely tailored to the diverse needs of various subjects.

\subsection{Conditioning Integration}

Integrating visual disparity-aware conditioning, text conditioning, and subject-augmented conditioning enables the personalized generation of dual-person portraits by leveraging the unique strengths of each modality. This holistic approach harnesses the detailed nuances of visual elements and descriptive text to ensure that the generated images reflect the specific characteristics and contexts of the subjects involved.

In DHumanDiff, a larger CLIP text encoder, $\Phi_2$, provides text conditioning, which is defined as $\mathbf{c}_T = \Phi_2(P)$. The cross-attention layer of the U-Net was adjusted to manage both text- and subject-augmented conditioning to produce text-oriented subject-augmented attention:
\begin{equation} \small
\mathbf{A}_{TS} = \text{Attention}(\mathbf{Q}, \mathbf{K}_{TS}, \mathbf{V}_{TS}) = \text{Softmax}\left(\frac{\mathbf{Q}(\mathbf{K}_{TS})^T}{\sqrt{d}}\right)\mathbf{V}_{TS}
\end{equation}
where $\mathbf{A}_{TS}$ is attention output, the query matrix $\mathbf{Q} = \mathbf{z}_t \mathbf{W}_q$ corresponds to the key matrices $\mathbf{K}_{TS} = (\mathbf{c}_T \, || \, \mathbf{c}_S) \mathbf{W}_{k_{TS}}$ and the value matrices $\mathbf{V}_{TS} = (\mathbf{c}_T \, || \, \mathbf{c}_S) \mathbf{W}_{v_{TS}}$, and $\mathbf{W}_{k_{TS}}$ and $\mathbf{W}_{v_{TS}}$ are the respective weight matrices. The final cross-attention formulation is expressed as follows:
\begin{equation} \small
\mathbf{A}_{\text{out}} = \mathbf{A}_{I_1} + \mathbf{A}_{I_2} +  \mathbf{A}_{TS}
\end{equation} 

To enhance text-to-image diffusion generation, we employ image prompt adapter technology. During training, the original U-Net parameters and all encoders are kept frozen, with updates applied exclusively to two additional image cross-attention layers. This strategy significantly reduces computational costs while maintaining the generalizability of the image generation model. Additionally, the training of DHumanDiff also includes an MLP and two transformer projectors. The DHumanDiff model is optimized by minimizing the following loss function:
\begin{equation} \small
\mathcal{L}_{\text{noise}}(\theta) = \mathbb{E}_{\mathbf{z}, \mathbf{c}, \boldsymbol{\epsilon} \sim \mathcal{N}(0,1), t} \left[ \left\| \boldsymbol{\epsilon} - \boldsymbol{\epsilon}_\theta(\mathbf{z}_t, t, \mathbf{c}) \right\|_2^2 \right]
\end{equation}
where $\left\| \boldsymbol{\epsilon} - \boldsymbol{\epsilon}_\theta(\mathbf{z}_t, t, \mathbf{c}) \right\|_2^2$ represents the squared Euclidean distance between the predicted and actual noise and where $\mathbb{E}$ denotes the expected value operator.

\subsection{Cascaded Inference Mechanism} 
As the reference inputs are facial images, directly applying image conditioning may lead to an overemphasis on facial regions, thereby compromising the generation of customized elements such as the background scene, clothing, and action details.

To balance the influence of image and text conditioning during generation, we introduce a cascaded inference mechanism. In the early denoising stages ($t \leq t_1$), the image conditioning weight $\lambda$ is set to a value less than 1, thereby emphasizing text-driven guidance. The guided noise prediction $\boldsymbol{\epsilon}'_t$ is then computed as:
\begin{equation} \small 
\hat{\boldsymbol{\epsilon}}'_t = \boldsymbol{\epsilon}_\theta(\mathbf{z}_t, t, \emptyset) + w \cdot \left( \boldsymbol{\epsilon}_\theta(\mathbf{z}_t, t, \mathbf{c}_{\lambda}) - \boldsymbol{\epsilon}_\theta(\mathbf{z}_t, t, \emptyset) \right)
\end{equation}
where $\boldsymbol{\epsilon}_\theta(\mathbf{z}_t, t, \mathbf{c}_{\lambda})$ represents the conditional noise prediction modulated by the image conditioning weight $\lambda$. The corresponding cross-attention output is computed as:
\begin{equation} \small \mathbf{A}'_{out} = \lambda \cdot (\mathbf{A}_{I_1} + \mathbf{A}_{I_2}) + \mathbf{A}_{TS} \end{equation} 

Subsequently, an attention map $\mathbf{M}$ (for the first token embeddings in $\mathbf{c}_{I_1}$ and $\mathbf{c}_{I_2}$) is computed to guide the noise fusion process:
\begin{equation} \small
\mathbf{M} = \text{Softmax}\left(\frac{\mathbf{Q}(\mathbf{K}_{I_1})^T}{\sqrt{d}}\right) + \text{Softmax}\left(\frac{\mathbf{Q}(\mathbf{K}_{I_2})^T}{\sqrt{d}}\right)
\end{equation} 

When $t>t_1$, the noise prediction $\boldsymbol{\epsilon}''_t$ is computed via a fusion strategy modulated by $\mathbf{M}$:
\begin{equation} 
\begin{aligned}
\small
\hat{\boldsymbol{\epsilon}}''_t = 
(1 - \mathbf{M}) \cdot \left( \boldsymbol{\epsilon}_\theta(\mathbf{z}_t, t, \emptyset) 
+ w \cdot \left( \boldsymbol{\epsilon}_\theta(\mathbf{z}_t, t, \mathbf{c}_{\lambda}) 
- \boldsymbol{\epsilon}_\theta(\mathbf{z}_t, t, \emptyset) \right) \right) \\[4pt]
\quad + \mathbf{M} \cdot \left( \boldsymbol{\epsilon}_\theta(\mathbf{z}_t, t, \emptyset) 
+ w \cdot \left( \boldsymbol{\epsilon}_\theta(\mathbf{z}_t, t, \mathbf{c}) 
- \boldsymbol{\epsilon}_\theta(\mathbf{z}_t, t, \emptyset) \right) \right)
\end{aligned}
\end{equation}

The $t_1$ is the critical timestep at which the image layout is established. By adjusting $\lambda$, the model balances facial generation with customized elements to allow for flexible image composition.

\section{Experiments}

\subsection{Experimental Setup}
\subsubsection{Datasets Overview}
We evaluate our method on three datasets:
\textbf{(i)FFHQ-wild Dataset.} The FFHQ-wild dataset \cite{Karras2021FFHQ} (hereafter referred to as FFHQ) consists of 70K high-resolution portrait images, encompassing both single-subject and multi-subject compositions with a majority of half-body shots. 
\textbf{(ii)PairHuman Dataset.} The PairHuman dataset contains 100K dual-person portraits. The dataset is split using stratified random sampling at a 9:1 ratio across three dimensions: four portrait topics, three composition types, and diverse scene categories (refer to Section 3.4.1). During preprocessing for training, all images were randomly cropped and resized to $512\times512$ pixels. 
\textbf{(iii)External Test Set.} We constructed an external test set with 21 identities, covering diverse ethnicities, skin tones, and styles. For each identity, five images are used for fine-tuning, and one representative image is selected for tuning-free inference. In addition, we wrote 20 descriptive prompts that cover the variations in clothing, scene, and pose. For quantitative assessment, we formed 12 distinct dual identity pairs and sampled each pair-prompt combination three times.

\subsubsection{Implementation Details}
The training of DHumanDiff was initiated based on the SDXL model \cite{Podell2023SDXL}, and all training images were resized to a resolution of 512$\times$512 pixels. To encode reference facial images, we used CLIP-ViT-Large-Patch14 \cite{alec2021clip} as the image encoder and the InsightFace VGG model \cite{deng2019arcface} as the face encoder. During training, all encoders and the original weights of the U-Net model were frozen, and only the newly added image cross-attention layers, MLP, and transformer projectors were trained. 

The model was trained for 20K steps on four NVIDIA A800 GPUs with a batch size of 64, using mixed-precision training with bfloat16 (bf16). We employed the AdamW optimizer with a learning rate of 1e-5 and a weight decay of 0.01. The implementation was based on PyTorch, running on Python 3.10 and CUDA 12.1. During inference, we used a DDIM sampler with 50 denoising steps and set the classifier-free guidance scale to 7.5.

\subsubsection{Evaluation Metrics}
Five key metrics were used to assess the generated image results. Face similarity was measured using MTCNN \cite{zhang2016MTCNN} and FaceNet \cite{Schroff2015facenet} for accurate facial feature replication. CLIP-T was used to assess the consistency between generated images and text prompts to ensure that the images accurately reflected the described scenarios. CLIP-I was used to assess the alignment of generated images with original dual-person images to ensure image authenticity \cite{alec2021clip}. PickScore \cite{kirstain2023pickscore} was applied to gauge user satisfaction by analyzing alignment with subjective preferences. Finally, MPS \cite{zhang2024MPS} provided a comprehensive assessment of human preferences, image detail, and semantic alignment to ensure overall image quality.

\subsection{Comparison with multi-subject personalization methods}

\subsubsection{Quantitative Evaluation}

The compared methods were Text Inversion \cite{gal2023TextualInversion}, CustomDiffusion \cite{Kumari2022Custom}, FastComposer \cite{xiao2024fastcomposer}, FaceDiffuser \cite{wang2024facediffuser}, and IP-Adapter \cite{Ye2023IPAdapter}. All models were trained at $512\times512$ resolution. Among these, Text Inversion and CustomDiffusion require individual-specific fine-tuning. FastComposer and FaceDiffuser were pretrained on the FFHQ dataset\cite{Karras2021FFHQ}. Additionally, IP-Adapter was enhanced by integrating two additional layers of cross-attention to enable the generation of two people. Quantitative results on the test subset of the PairHuman dataset and the external dataset are shown in Table~\ref{tab:result_comparison} and Table~\ref{tab:result_comparison_new}, respectively. The comparative analysis is detailed as follows.

\begin{table}[ht]
\centering
\caption{Quantitative comparison with existing multi-subject personalization methods on the PairHuman test subset.}
\renewcommand{\tabcolsep}{2.5pt} 
\label{tab:result_comparison}
\scriptsize
\begin{tabular}{ccccccc}
\hline
\textbf{Type} & \textbf{Methods} & \textbf{Face Sim.} & \textbf{CLIP-T} & \textbf{CLIP-I} & \textbf{PickScore} & \textbf{MPS} \\
\hline
\multirow{2}{*}{FFHQ} & FastComposer & 0.6199 & 17.0983 & 0.7787 & 21.1478 & 8.6704 \\
 & Face-Diffuser & 0.6059 & 17.1053 & 0.7707 & 21.1575 & 8.7246 \\ \hline
\multirow{4}{*}{PairHuman} & FastComposer & 0.5962 & 26.5453 & 0.8232 & 21.7239 & 8.7829 \\
 & Face-Diffuser & 0.5557 & \textbf{26.5662} & 0.8159 & 21.5650 & 8.5023 \\
 & IP-Adapter & 0.6028 & 26.2356 & 0.8225 & 21.6859 & 9.1283 \\
 & DHumanDiff & \textbf{0.6232} & 26.2862 & \textbf{0.8377} & \textbf{21.8273} & \textbf{9.1391} \\ 
\hline
\vspace{-1em}
\end{tabular}
\end{table}

\begin{table}[ht]
\centering
\caption{Quantitative comparison with existing multi-subject personalization methods on the external test dataset.}
\renewcommand{\tabcolsep}{2.5pt} 
\label{tab:result_comparison_new}
\scriptsize
\begin{tabular}{ccccccc}
\hline 
\textbf{Type} & \textbf{Methods} & \textbf{Face Sim.} & \textbf{CLIP-T} & \textbf{PickScore} & \textbf{MPS} \\ \hline
\multirow{2}{*}{Fine-tuning} & Text Inversion & 0.3351 & 25.4121 & 20.8433 & 8.5751 \\
 & CustomDiffusion & 0.3974 & 22.9241 & 20.8155 & 9.2461 \\ \hline
\multirow{2}{*}{FFHQ} & FastComposer & 0.6414 & 18.8813 & 19.7507 & 8.3867 \\
 & Face-Diffuser & 0.6277 & 20.2091 & 20.1890 & 8.6112 \\ \hline
\multirow{4}{*}{PairHuman} & FastComposer & 0.5229 & 25.9750 & 21.3544 & 9.4880 \\
 & Face-Diffuser & 0.4936 & 26.1940 & 21.1810 & 9.0372 \\
 & IP-Adapter & 0.6839 & 23.7725 & 21.3393 & 10.0840 \\
 & DHumanDiff & \textbf{0.7010} & \textbf{26.2474} & \textbf{22.2044} & \textbf{10.8543} \\ \hline
\end{tabular}
\end{table}

\textbf{Dataset Comparison.} The results in the Table~\ref{tab:result_comparison} and Table~\ref{tab:result_comparison_new} reveal significant performance variations between models trained on the FFHQ and PairHuman datasets. Specifically, FastComposer and FaceDiffuser trained on the PairHuman dataset exhibited lower face similarity than those trained on the FFHQ dataset, with FaceDiffuser showing a more pronounced decrease. This decline was primarily attributed to the smaller facial sizes in the PairHuman dataset and FaceDiffuser’s emphasis on textual consistency, which compromises facial detail. Despite this, methods trained on PairHuman exhibited improvements in the CLIP-T, CLIP-I, and PickScore metrics, indicating that the high-quality images and detailed annotations of the PairHuman dataset enhance customized generation and visual quality.

\textbf{Method Comparison.} 
As shown in Table~\ref{tab:result_comparison}, among the PairHuman-trained methods, FaceDiffuser and DHumanDiff achieved higher CLIP-T scores, demonstrating strong textual consistency. IP-Adapter and DHumanDiff achieved higher face similarity and MPS scores, with DHumanDiff outperforming all other methods in nearly every metric.
As shown in Table~\ref{tab:result_comparison_new}, fine-tuning-based methods generally perform worse, particularly in preserving identity fidelity. This limitation arises from Textual Inversion’s inability to handle multiple identities simultaneously and CustomDiffusion’s difficulty with multiple instances of the same subject class. Among all the methods compared, DHumanDiff outperforms in all evaluation metrics. 
Notably, the higher scores observed on the external test set can be attributed to its use of high-resolution frontal-face images, which are more conducive to identity preservation compared to the diverse and complex samples in the PairHuman test set.

\begin{figure*}[!t]
\newcommand{\uwidth}{5.5} 
\newcommand{\uhoriz}{0.07}  
\small
\centering{ %
\begin{minipage}{ 0.97 \textwidth} %
\centering{
\includegraphics[width= \uwidth in]{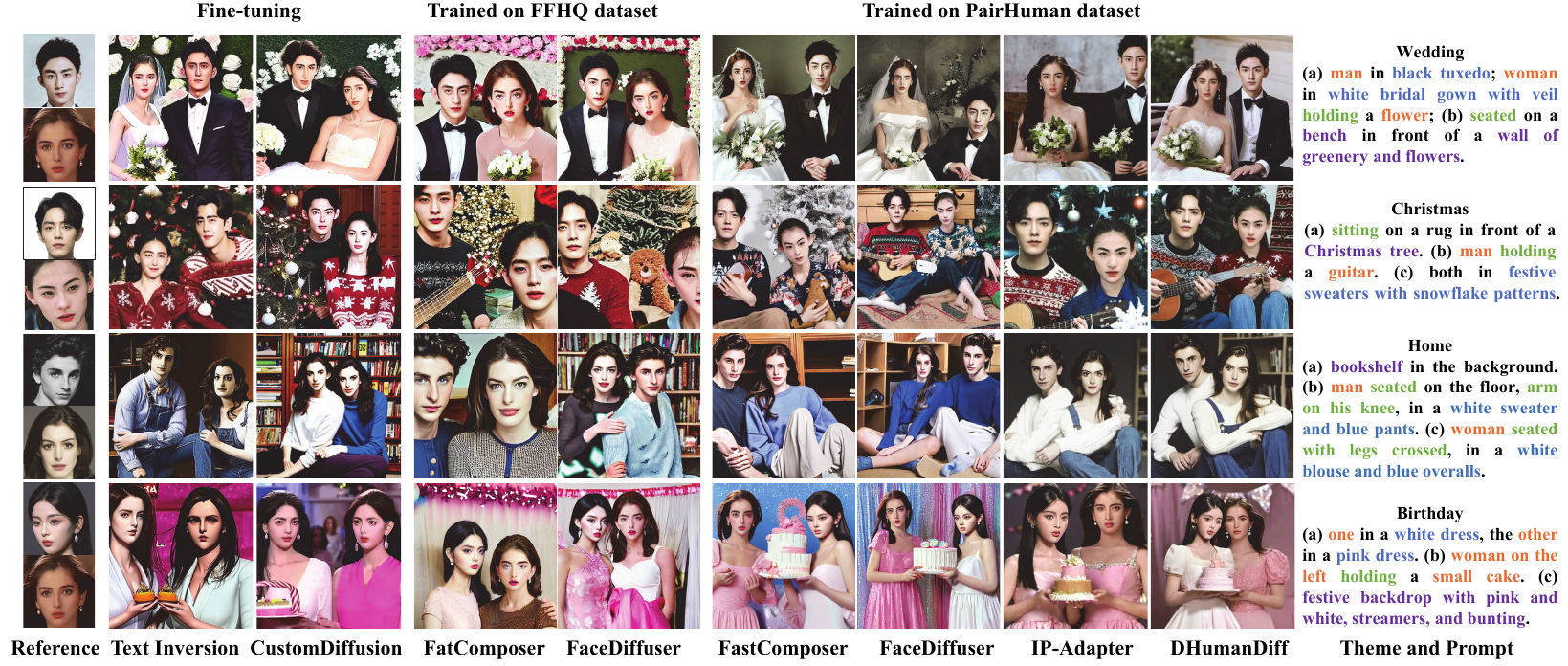}
}
\end{minipage}
}
\vspace{-1 em} 
\caption[Comparison]{Comparison of fine-tuning methods, FFHQ-trained methods, and PairHuman-trained methods, showcasing dyadic portraits across four themes: wedding, Christmas, home, and birthday.\footnotemark} 
\vspace{-1 em} 
\label{fig:comparison_results}
\end{figure*}

\footnotetext{The exhaustive text prompts is available in the \ref{appendix: Prompts for Dual-person Portrait Genertation}.}

\subsubsection{Qualitative Evaluation}
\textbf{Comparison of Datasets.} As shown in Fig.~\ref{fig:comparison_results}, the PairHuman-trained methods (FastComposer/ FaceDiffuser) outperformed the FFHQ-trained models in generating customized dual-person portraits across three key aspects: \textbf{(i) Specific detail generation.} The PairHuman-trained methods accurately generated specific elements (e.g., ``white bridal veil"' in the first row and ``cake'' in the fourth row) and matched colors (e.g., ``white dress'' and ``pink dress'' in the fourth row), whereas the FFHQ-trained models struggled with such details. \textbf{(ii) Pose accuracy.} The PairHuman-trained methods generated complex human poses with higher precision, such as ``man seated on the floor with an arm on his knee'' and ``woman seated with legs crossed'' in the third row. \textbf{(iii) Composition and visual appeal.} Unlike the FFHQ-trained methods, the PairHuman-trained methods handled complex scenes with multiple elements (e.g., Christmas trees, violins, and two individuals in the second row) more effectively, thus producing well-composed, visually appealing portraits. In summary, the rich visual content and detailed annotations in the PairHuman dataset enable highly customized dual-person portrait generation to address a wide range of user needs.

\textbf{Comparison of the Methods.} As shown in Fig.~\ref{fig:comparison_results}, the fine-tuning methods and PairHuman-trained methods differed in several areas: \textbf{(i) Facial consistency.} The fine-tuning methods (Text Inversion and CustomDiffusion) often exhibited identity confusion. In contrast, PairHuman-trained methods maintained consistent facial features, with DHumanDiff performing best. \textbf{(ii) Image layout.} FaceDiffuser provided greater layout variability but sacrificed facial fidelity, especially in distant shots such as the third row. The other methods prioritized subject prominence, leading to smaller background sizes, where IP-Adapter produced the largest subjects. \textbf{(iii) Textual consistency.} Text Inversion and CustomDiffusion failed to generate specific elements, such as ``guitar'' in the second row and ``woman seated with legs crossed'' in the third row. IP-Adapter also struggled with generating detailed attributes such as ``a man's arm on his knee" in the third row and ``white dress" in the fourth row. In contrast, FastComposer, FaceDiffuser, and DHumanDiff demonstrated better adherence to textual prompts. Overall, DHumanDiff achieves superior facial consistency and effective customization in clothing, actions, and scenes.

\begin{table}[htbp]
\centering
\caption{Comparison of training resources and efficiency of different Methods.}
\renewcommand{\tabcolsep}{2.5pt} 
\scriptsize
\label{tab:train_gpu_hours}
\centering
\begin{tabular}{cccccc}
\hline
\multirow{2}{*}{\textbf{Methods}} & \textbf{Fine-} & \textbf{\#Image} & \textbf{Training} & \textbf{GPU}  & \textbf{Training}  \\
 & \textbf{Tuning} & \textbf{Per Subject} & \textbf{Step} & \textbf{hours} & \textbf{Parameter Size} \\
\hline
Text Inversion  & Y & 5 & - & - & -\\
CustomDiffusion  & Y & 5 & - & - & -\\
FastComposer & N & 1  & 100K & 2096h & 6G\\
FaceDiffuser & N & 1 & 100K & 4256h & 12G\\
IP-Adapter & N & 1 & 10K & 16h & 0.28G\\
DHumanDiff & N & 1  & 20K & 32h & 0.3G\\
\hline
\end{tabular}
\end{table}

\subsubsection{Complexity Analysis}
Table~\ref{tab:train_gpu_hours} compares the training resources and efficiency of the methods. Text Inversion and CustomDiffusion require fine-tuning for each subject, utilizing at least five images per subject. In contrast, FastComposer, FaceDiffuser, IP-Adapter, and DHumanDiff enable direct inference from a single image per subject, which markedly enhances user convenience. These methods were trained on the PairHuman dataset, which contains 90K training samples. FastComposer and FaceDiffuser used eight NVIDIA A800 GPUs with a batch size of 128 to complete 100K training steps. These methods need updates across the entire U-Net architecture, with FaceDiffuser training two models, which resulted in total training parameter sizes of 6 GB and 12 GB. Consequently, they required substantial training times of 2,096 and 4,256 GPU hours, respectively. In contrast, IP-Adapter and DHumanDiff froze the original U-Net model and required only approximately 0.3 GB of training parameters, which significantly enhanced the training efficiency. They were trained on four NVIDIA A800 GPUs for 10K and 20K steps and consumed just 16 and 32 GPU hours, respectively.

\subsection{Comparison with SOTA single-subject personalization methods}

\begin{table}[ht]
\centering
\caption{Quantitative comparison with single-subject personalization methods on the external test dataset.}
\renewcommand{\tabcolsep}{2.5pt} 
\label{tab:result_comparison_single}
\scriptsize
\begin{tabular}{cccccc}
\hline 
\textbf{Methods} & \textbf{Face Sim.} & \textbf{CLIP-T} & \textbf{PickScore} & \textbf{MPS} \\ \hline
InstantID (512$\times$512) & 0.5265 & 26.0890 & 21.0682 & 9.0889 \\
InstantID (1024$\times$1024) & 0.6881 & 27.0604 & 22.0677 & 9.4001 \\
PuLID (512$\times$512) & 0.4625 & 26.3192 & 22.0997 & 10.1383 \\
PuLID (1024$\times$1024) & 0.6295 & \textbf{27.1375} & 22.1358 & 10.4143 \\
DHumanDiff (512$\times$512) & \textbf{0.7010} & 26.2474 & \textbf{22.2044} & \textbf{10.8543} \\ \hline 
\end{tabular}
\end{table}

To enable personalized generation of two persons, InstantID \cite{Wang2024InstantID} and PuLID \cite{Guo2024PuLID} were integrated into the ComfyUI framework. Both workflows employ the attention mask node for spatial control, while ControlNet-Pose guidance is applied to the InstantID workflow. Due to the difference in training resolutions (InstantID and PuLID were trained at $1024\times1024$, while DHumanDiff was trained at $512\times512$), we report results at both $512\times512$ and $1024\times1024$. 

\textbf{Quantitative Comparison}. All quantitative metrics were computed on the external test set because the evaluated workflows require either subject-specific prompt separation or pose-map selection. As shown in Table~\ref{tab:result_comparison_single}, InstantID and PuLID achieve higher CLIP-T scores, whereas DHumanDiff demonstrates superior performance in face similarity, MPS, and PickScore. The improvement in face similarity is particularly evident at $512\times512$. Notably, both InstantID and PuLID show a decline in facial fidelity compared to their performance on single-subject inputs. Overall, DHumanDiff achieves satisfactory identity fidelity and visual quality at the current training resolution, while exhibiting potential for further improvement at higher resolutions.

\begin{figure*}[!t]
\newcommand{\uwidth}{5.5} 
\newcommand{\uhoriz}{0.07}  
\small
\centering{ %
\begin{minipage}{ 0.97 \textwidth} %
\centering{
\includegraphics[width= \uwidth in]{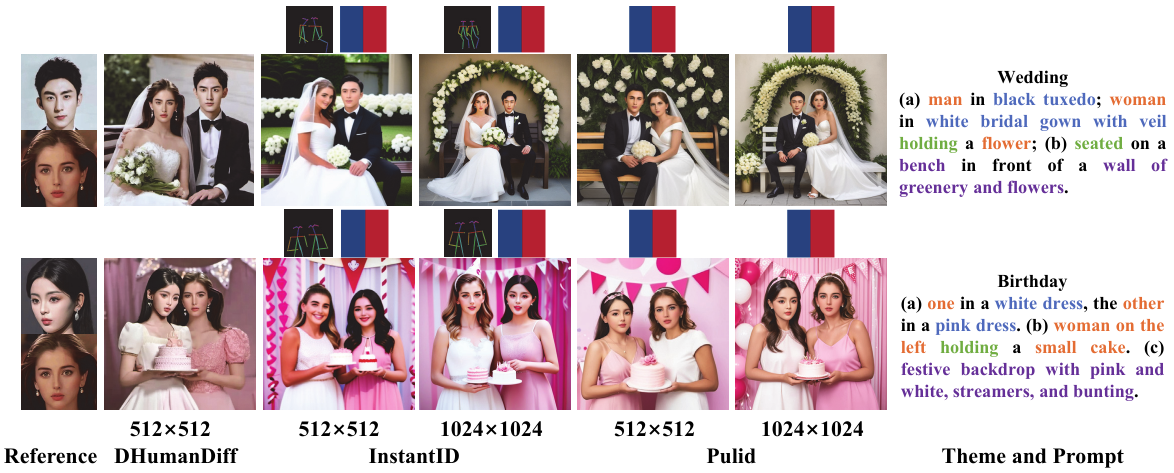}
}
\end{minipage}
}
\vspace{-1 em} 
\caption{Comparison with InstantID and PuLID models implemented on the ComfyUI platform. Both compared methods require layout conditions for attention mask isolation, with InstantID further requiring a pose reference for ControlNet.}
\vspace{-1 em} 
\label{fig:comparison_results_single}
\end{figure*}

\textbf{Qualitative Analysis}. As shown in Fig.~\ref{fig:comparison_results_single}, all models demonstrate good performance overall. PuLID produces a more balanced subject-background arrangement while retaining strong semantic alignment with the textual prompt. InstantID relies on external pose cues and requires precise pose configuration to generate dual-person compositions. While it generally adheres to the prompts, minor discrepancies in fine-grained details (e.g., cake counts) may still arise. In DHumanDiff, the subject is rendered more prominently, leaving limited space for background elements, which may compromise text-image consistency. This aspect could be further improved in future work. Regarding facial consistency, both InstantID and PuLID show a significant decrease at $512\times512$. Even at their native $1024\times1024$, their ability to preserve facial identity is still inferior to that of DHumanDiff.

\subsection{Ablation Study}

\begin{table}[ht]
\centering
\caption{Ablation analysis of the proposed method}
\label{tab:Ablation}
\renewcommand{\tabcolsep}{2.5pt} 
\scriptsize
\centering
\begin{tabular}{cccccc}
\hline
\textbf{Methods} & \textbf{Face Sim.} & \textbf{CLIP-T} & \textbf{CLIP-I} & \textbf{PickScore} & \textbf{MPS} \\ \hline
w/o visual disparity-aware conditioning & 0.5174 & \textbf{26.4071} & 0.8241 & 21.8003 & \textbf{9.4338} \\
w/o subject-augmented conditioning & 0.6028 & 26.2356 &	0.8225 &	21.6859	& 9.1283  \\ 
w/o ID enhancement module & 0.5996 &	26.3713	& 0.8222	& 21.6628 &	9.0405  \\
DHumanDiff &  \textbf{0.6232} & 26.2862 & \textbf{0.8377} & \textbf{21.8273} & 9.1391 \\ \hline
\vspace{-1em}
\end{tabular}
\end{table}

We conducted ablation experiments to assess the contribution of each key component in DHumanDiff, as shown in Table~\ref{tab:Ablation}. \textbf{(i) Visual disparity–aware conditioning.} The two-layer image cross-attention was replaced with a shared attention mechanism that shares key and value projections across reference images. This change led to a significant decrease in face similarity, confirming the need for visual disparity-aware conditioning to distinguish facial characteristics and generate high-fidelity dual-person portraits. \textbf{(ii) ID enhancement module.} Removal of this module reduced face similarity, despite a slight improvement in CLIP-T, confirming that the integration of local and global facial characteristics enhances the preservation of identity. \textbf{(iii) Subject-augmented conditioning.} Eliminating this component degrades all metrics, showing its effectiveness in improving both semantic alignment and facial fidelity by aligning text embeddings with image features. In summary, the proposed architecture substantially improves facial consistency, with only minor trade-offs in text-image alignment, enabling high-fidelity dual-person portrait generation.

\begin{figure}[!t]
\newcommand{\uwidth}{3.4}
\small
\centering{ %
\begin{minipage}{ 0.7 \textwidth} %
\centering{
\includegraphics[width= \uwidth in]{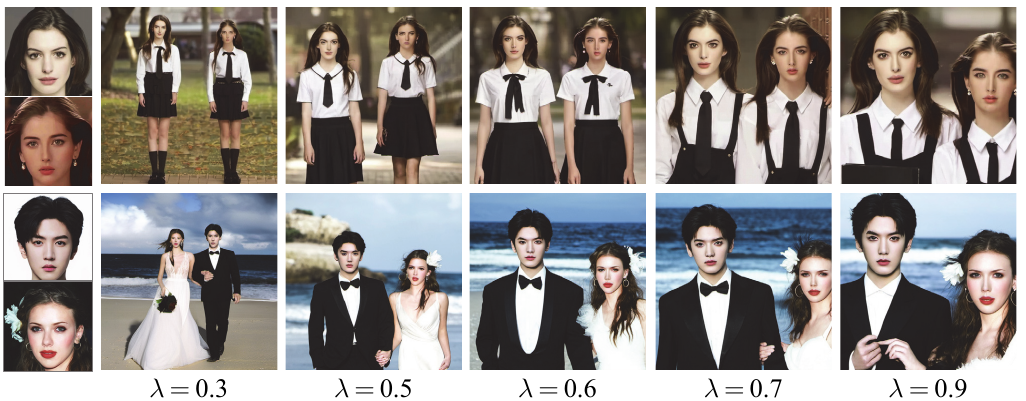}
}
\end{minipage}
}
\vspace{-0.5 em} 
\caption{Ablation studies on the hyper-parameter $\lambda$.}
\vspace{-0.5em} 
\label{fig:hyperparameter}
\end{figure}

\subsection{Hyperparameter Analysis} 
Fig.~\ref{fig:hyperparameter} shows the effects of varying $\lambda$ on the images generated for two text prompts: ``a woman and a woman stand in park, dressed in school uniforms of white shirts and black skirts'' and ``a man in a black suit and a woman in a white gown walking hand in hand on the beach.'' At $\lambda=0.3$, the images typically show full-body views with blurred facial details. As $\lambda$ increases, the proportion of faces in the images increases, and facial clarity improves. However, values above 0.7 may lead to images that focus excessively on the face and fail to capture the described actions and attire. To achieve a better balance between textual consistency, image layout diversity, and facial feature preservation, setting $\lambda$ within the range of 0.5 to 0.7 is recommended.

\subsection{Saliency Analysis}

\begin{figure}[!t]
\newcommand{\uwidth}{5}
\small
\centering{ %
\begin{minipage}{ \textwidth} %
\centering{
\includegraphics[width= \uwidth in]{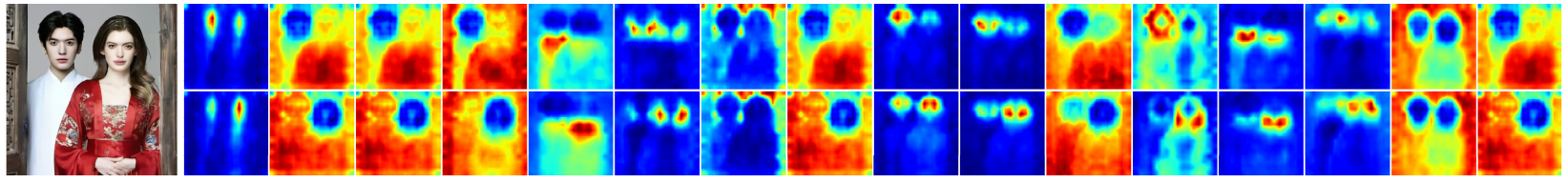}
}
\end{minipage}
}
\vspace{-0.5 em} 
\caption{Cross‐attention heatmaps for all 16 visual tokens of $\mathbf{c}_{I_1}$ (top row) and $\mathbf{c}_{I_2}$ (bottom row). Warmer colors indicate stronger attention.}
\vspace{-0.5em} 
\label{fig:heatmap}
\end{figure}

Fig.~\ref{fig:heatmap} displays the average cross-attention weights of all visual tokens for embeddings $\mathbf{c}_{I_1}$ (top row) and $\mathbf{c}_{I_2}$ (bottom row). For $\mathbf{c}_{I_1}$, attention is primarily focused on the facial region of subject 1 and its surrounding background, while $\mathbf{c}_{I_2}$ focuses predominantly on the facial area of subject 2 and the surrounding context. These distinct activation patterns indicate the effectiveness of visual-disparity conditioning in distinguishing identity-specific features and reducing feature leakage between subjects. As a result, DHumanDiff enables high-fidelity dual-person portrait generation while maintaining the distinct identities of both subjects.

\section{Limitations and Future Work}

\textbf{Demographic Diversity:} The PairHuman dataset was designed to improve representation of Asian identities in image generation, and DHumanDiff trained on it generalizes across age groups and ethnicities (see  \ref{age}). However, its current lack of demographic diversity may limit its applicability across broader cultural and aesthetic contexts. Future work could expand the dataset’s diversity by incorporating high-resolution imagery from more varied sources (e.g., media and everyday contexts). In addition, targeted data collection may help improve balance across age, ethnicity, and geography.

\textbf{Sensitivity to Lighting:} Our analysis reveals that DHumanDiff exhibits sensitivity to lighting conditions (see \ref{light}). One promising approach to improving robustness is training with lighting-augmented data. Additionally, incorporating 3D-aware modeling techniques to decouple illumination from identity may further enhance the model's ability to handle complex lighting environments.

\textbf{Scalability to Multi-Person Scenarios:} DHumanDiff currently supports only dual-subject generation due to its reliance on paired cross-attention mechanisms. Expanding the model to support multi-person scenarios will require the development of scalable tokenization strategies and adaptive attention mechanisms. Additionally, high-quality multi-subject datasets will be crucial for training and evaluating the model's performance in these more complex settings.

\textbf{Architectural Enhancements:} The DHumanDiff model has considerable potential for improvement in semantic consistency and image fidelity. Future research directions will investigate transformer-based diffusion architectures \cite{flux2024}, leveraging powerful language encoders (e.g., T5-XXL), and integrating advanced cross-modal attention mechanisms (e.g., MM-DiT) to enhance both alignment and generation quality.

\section{Conclusion}

In this study, we introduced the PairHuman dataset, a carefully curated collection of 100K high-resolution dual-person portraits that are notable for their photographic quality, diverse visual content, and rich annotations. In addition, we proposed the DHumanDiff method, which is a new approach for generating highly customized dual-person portraits with enhanced facial feature preservation and adjustable image layouts. Experimental results demonstrate that DHumanDiff achieves an improvement in face similarity. Furthermore, the PairHuman dataset shows notable improvements in CLIP-T and CLIP-I scores over the benchmark, affirming its effectiveness for high-fidelity customization. A key limitation is the dataset’s predominant focus on young Asian adults, which future work should address to enhance the inclusivity and applicability of the PairHuman dataset.

\section{Ethical Statement.} 

The PairHuman dataset was collected from publicly available web sources (Baidu), and all images were manually filtered to remove sensitive, copyrighted, or other inappropriate content. To mitigate privacy risks, all source metadata (EXIF, GPS/geolocation, and source/URL fields) were systematically stripped to prevent accidental privacy disclosure. Annotations include only general attributes (e.g., gender, age range, pose) and do not contain personally identifiable information. Additionally, a safety checker was integrated into the generation pipeline to filter inappropriate outputs.

In alignment with international privacy standards such as GDPR, we developed an anonymization pipeline that replaces all original facial identities with synthetically generated faces. This process thoroughly eliminates identifiable facial features while maintaining the utility of the data for model training. The publicly released PairHuman dataset consists exclusively of anonymized images. A verified subset of the anonymized dataset, along with the anonymization code, has been made openly accessible at: https://github.com/annaoooo/PairHuman.

\section{CRediT authorship contribution statement}
Ting Pan: Data curation, Methodology, Writing – original draft, Visualization, Validation;  
Ye Wang: Methodology, Writing – original draft, Validation; 
Peiguang Jing: Supervision, Writing – review and editing; 
Rui Ma: Investigation, Writing – review and editing; 
Zili Yi: Supervision, Conceptualization, Funding acquisition; 
Yu Liu: Supervision, Conceptualization, Funding acquisition.

\section{Declaration of competing interest}
The authors declare that they have no known competing financial interests or personal relationships that could have appeared to influence the work reported in this article.

\section{Acknowledgments}
This work is supported by the National Natural Science Foundation of China under Grant (Grant No. 61771338) and the Young Scientists Fund of the National Natural Science Foundation of China under Grant (Grant No.   ).

\section{Data availability}
The PairHuman datasets are publicly available and are provided in the links within the article.

\appendix
\setcounter{table}{0}
\setcounter{figure}{0}
\setcounter{algorithm}{0}

\section{Expanded Experimental Analysis}

\subsection{Generalization Across Age and Ethnic Diversity}

\label{age}

\begin{figure}[!t]
\newcommand{\uwidth}{3}
\small
\centering{ %
\begin{minipage}{ \textwidth} %
\centering{
\includegraphics[width= \uwidth in]{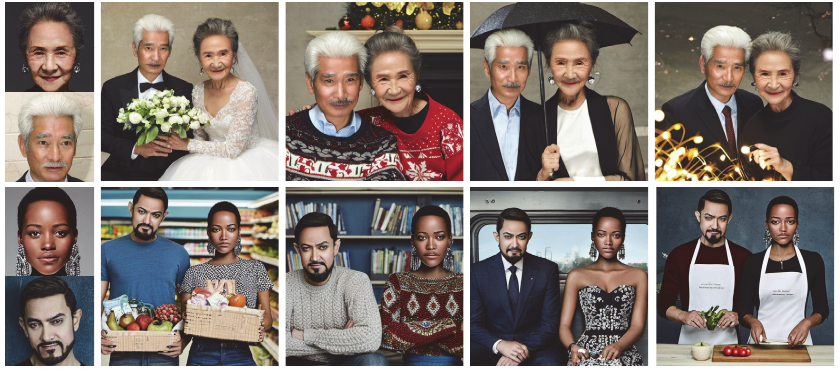}
}
\end{minipage}
}
\vspace{-0.5 em} 
\caption{Generalization analysis of DHumanDiff across age and ethnic diversity}
\vspace{-0.5em} 
\label{fig:generalization}
\end{figure}

To evaluate the generalization performance of DHumanDiff under demographic variations, we conducted additional experiments on two representative reference pairs: (i) two elderly individuals and (ii) two subjects exhibiting significant differences in skin tone and facial features compared to those presented in the main manuscript. As illustrated in Fig.~\ref{fig:generalization}, the proposed method successfully preserves identity consistency in both cases, indicating its ability to generate personalized dual-person portraits across diverse age groups and ethnic backgrounds. It is important to note that these demographic groups are not present in the current PairHuman dataset. We attribute DHumanDiff’s observed generalization capability to its robust ability to preserve facial features and skin texture consistency even under out-of-distribution conditions.

\subsection{Sensitivity Analysis of Reference Illumination and Quality}
\label{light}

\begin{figure}[!t]
\newcommand{\uwidth}{3.2}
\small
\centering{ %
\begin{minipage}{ \textwidth} %
\centering{
\includegraphics[width= \uwidth in]{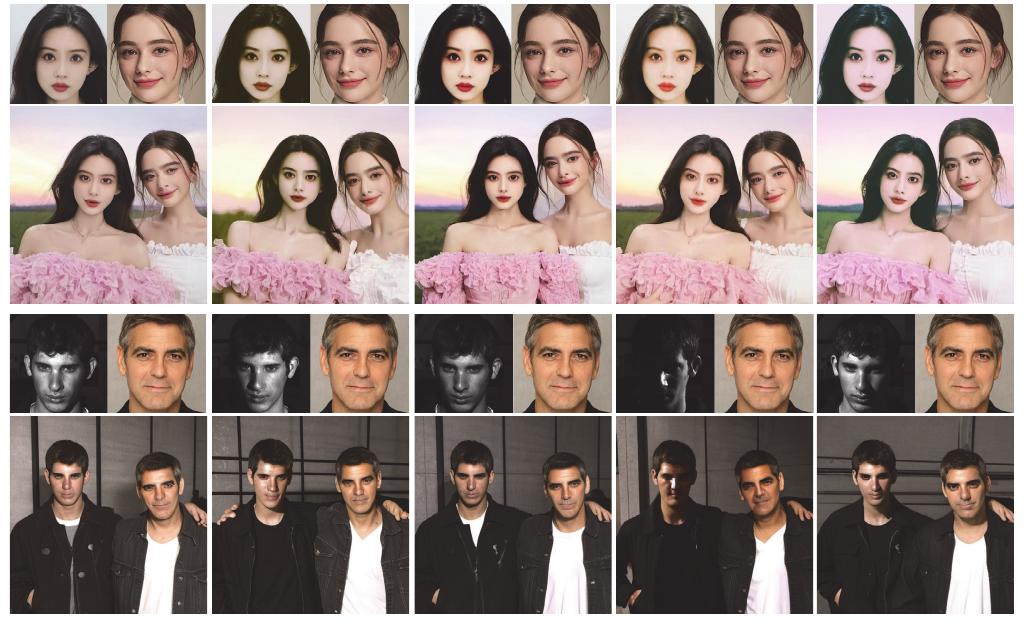}
}
\end{minipage}
}
\vspace{-0.5 em} 
\caption{Sensitivity analysis of the DHumanDiff model to illumination and quality of reference images.}
\vspace{-0.5em} 
\label{fig:illumination_sensitivity}
\end{figure}

As illustrated in Fig.~\ref{fig:illumination_sensitivity}, the DHumanDiff dual-person portrait generation model demonstrates sensitivity to both illumination conditions and the overall quality of the reference images. In the first example, unnatural lighting in the reference images may disrupt the global visual coherence of the generated portraits. The second example reveals that references with shadows or uneven illumination often produce undesirable shading artifacts, whereas low-visibility references result in a significant decline in overall image quality. Therefore, improving the model's robustness to such input variations remains a critical direction for future work.

\subsection{Diverse Pose and Layout Generation}
\label{pose}

\begin{figure}[!t]
\newcommand{\uwidth}{4}
\small
\centering{ %
\begin{minipage}{ \textwidth} %
\centering{
\includegraphics[width= \uwidth in]{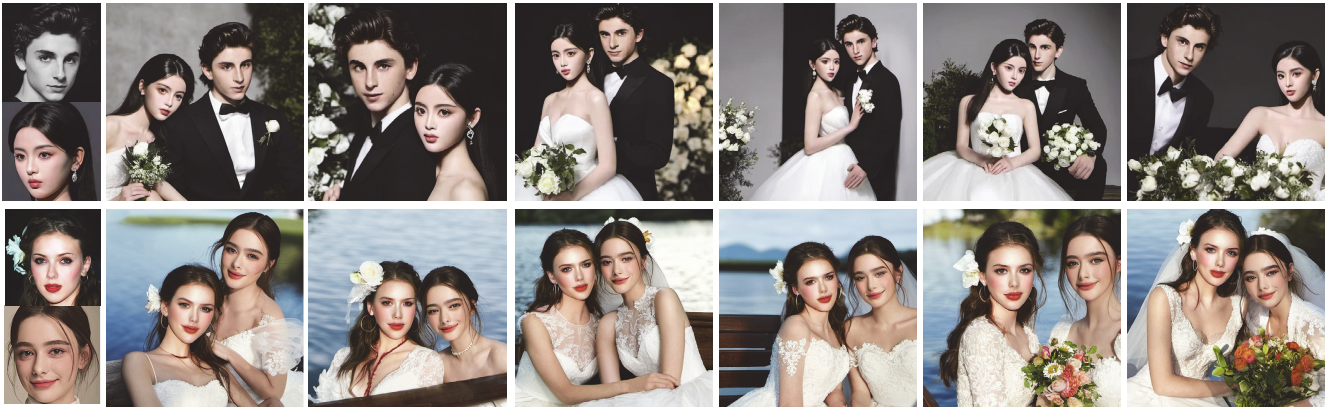}
}
\end{minipage}
}
\vspace{-0.5 em} 
\caption{Pose and layout diversity in dual-person portrait generation using DHumanDiff.}
\vspace{-0.5em} 
\label{fig:pose_diversity}
\end{figure}

To evaluate DHumanDiff’s capability to generate diverse human poses and spatial layouts, we conducted a supplementary experiment using two prompts: (i) ``A man and a woman pose for a formal portrait, with greenery and flowers adorning the background. The man is dressed in a black tuxedo and the woman is in a white wedding gown.'' (ii) ``A woman and a woman in white wedding dresses sit on a wooden bench by a serene body of water. Their dresses are ornate with intricate lace.'' For each prompt, we generated multiple outputs by varying the random seed. As shown in Fig.~\ref{fig:pose_diversity}, DHumanDiff generates a wide variety of subject compositions, body postures, and inter-subject interactions. Unlike methods relying on explicit spatial or pose conditioning, our approach enables diverse spatial layouts with greater simplicity and flexibility in implementation.

\subsection{User Study}

\begin{figure}[!t]
\newcommand{\uwidth}{5}
\small
\centering{ %
\begin{minipage}{ \textwidth} %
\centering{
\includegraphics[width= \uwidth in]{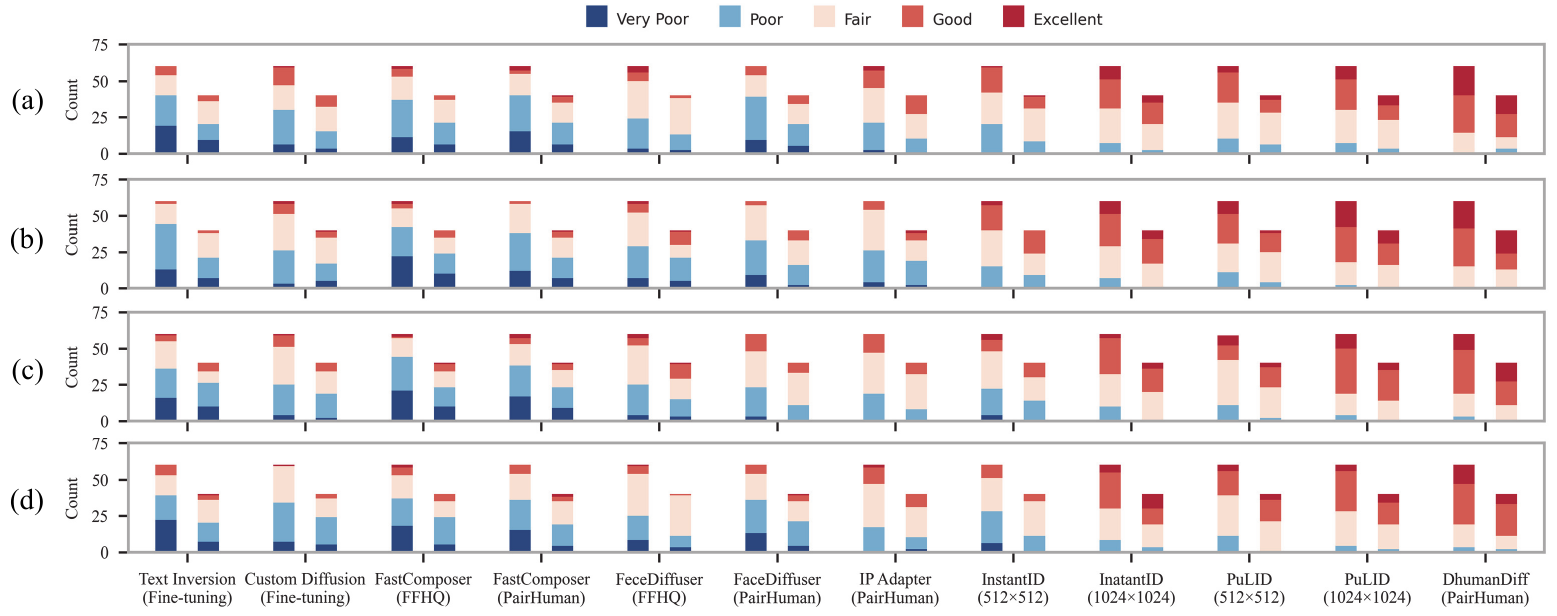}
}
\end{minipage}
}
\vspace{-0.5 em} 
\caption{Distribution of user ratings across evaluation criteria using a 5-point Likert scale (1: Very Poor, 2: Poor, 3: Fair, 4: Good, 5: Excellent), grouped by participant age. For each method, the left bar represents ratings from younger participants (18-39 years), and the right bar represents ratings from middle-aged participants (40-70 years). The results are shown for (a) facial similarity, (b) consistency with prompts, (c) portrait realism, and (d) willingness to adopt.}
\vspace{-0.5em} 
\label{fig:user_study}
\end{figure}

We conducted a user study to evaluate the quality and practical utility of our personalized dual-person portrait generation model. The study involved 100 participants (53 men, 47 women), with ages ranging from 18 to 70 years (mean=37.9, SD=11.7). Participants were recruited from diverse backgrounds, including 35 experts in computer graphics or generative AI and 65 non-experts.

We evaluated 12 methods, each producing three outputs from the same three reference face pairs and prompts. For each method, participants viewed its three generated images and provided one aggregate rating for each of the four evaluation criteria: (i) facial similarity to reference subjects, (ii) consistency with textual prompts, (iii) portrait realism, and (iv) willingness to use the generated images. To mitigate order effects, the presentation order of different methods was randomized for each participant. Each participant rated all methods on a 5-point Likert scale, yielding 4,800 total ratings (100 participants $\times$ 12 methods $\times$ 4 criteria).

Considering the wide age range of participants, we stratified them into two cohorts: younger participants (18-39 years) and middle-aged participants (40-70 years) for analysis.  Fig.~\ref{fig:user_study}, the results did not reveal substantial differences between age groups. In addition, statistical analyses across gender and professional expertise revealed no significant differences. Overall, DHumanDiff consistently received a higher proportion of favorable ratings (scores 4-5) across all evaluation criteria, and the aggregated results indicate that most participants preferred DHumanDiff for generating personalized dual-person portraits, although individual preferences varied.

\section{Data Collection Keywords}
\label{appendix: Data Collection Keywords}

The following presents all keywords used to guide the data collection process for the PairHuman dataset.

\begin{itemize}
\item{Cultural aesthetics include regional styles such as Hong Kong, Japanese, and Korean, as well as broader stylistic variations such as sophistication, woodland, freshness, minimalism, magazine style, creativity, fantasy, Instagram-inspired, vintage, artistry, nostalgia, nature, monochrome, vivacity, mood, and cinematic.}
\item{Emotional expressions include tenderness, romance, warmth, and sweetness.}
\item{Environmental settings span a variety of locations, including campuses, homes, grasslands, forests, beaches, streets, travel destinations, amusement parks, nighttime scenes, urban landscapes, and festive occasions such as Christmas, New Year, and birthdays.}
\item{Costume-specific themes encompass traditional Chinese garments, cheongsams, modern fashion, Hanfu, formal wear, ethnic attire, and national styles.}
\end{itemize}

\section{Prompts of LLaVA annotation}
\label{appendix: Prompts of LLaVA annotation}

We designed three prompts for LLAVA to generate annotations, which include image captions, photo themes, and bounding boxes (BBoxes) with attributes for persons and objects. The annotation prompts are as follows:

\begin{itemize}
  \item \textbf{Image Caption and Photo Theme:} Describe the dual-person photo, providing a brief image description, between 60 to 100 words, capturing all discernible elements, their attributes, locations, interactions, and the image's spatial layout. This caption will be used for image re-creation, so the closer the resemblance to the original image, the better the tag quality. Use words or phrases to detail the setting (indoor/outdoor), location, season, weather, festival, mood, photo style, clothing style, activity, and textual content.
  \item \textbf{Attributes of Person1 and Person2:} Describe separately all humans in the dual-person photo, providing details such as 'object\_id' (person1/person2), 'character angle' (frontal, side), 'gender', 'age group', and 'unique coordinate' (formatted as [x0, y0, x1, y1] for the person), 'facial traits', 'hair detail', 'skin tone', 'action' (activity or stance), 'expression', 'details of clothes and shoes', and 'accessories' (any additional items they are wearing or carrying). Ensure the description is succinct and precise.
  \item \textbf{Standout Non-human Items:} Identify and describe up to three standout non-human items in the dual-person photo, amalgamating similar items into collective descriptions. Explore the interaction between these objects and the people. Provide details like 'object\_id' (the category label), 'count', 'coordinates' (formatted as [x0, y0, x1, y1] for each item), 'appearance', and 'interaction' (brief description of both parties and their relationship). Ensure the description is unique, concise, and accurate.
\end{itemize}

\section{Prompts for Dual-person Portrait Genertation}
\label{appendix: Prompts for Dual-person Portrait Genertation}

The complete text prompts corresponding to the four themes in the qualitative comparative experiment are as follows:
\begin{itemize}
\item{\textbf{Wedding:} A man and a woman pose for a formal portrait. The man is dressed in a black tuxedo. The woman is in a white bridal gown with a veil, holding a bouquet of white flowers. A man and a woman are seated on a bench in front of a wall adorned with greenery and flowers.}
\item{\textbf{Christmas:} A man and a woman, sitting on a rug in front of a christmas tree. The man is holding a guitar. A woman and a man are both wearing festive sweaters with snowflake patterns. The christmas tree is adorned with ornaments and a star on top.}
\item{\textbf{Home:} A man and a woman pose for a photo, with a bookshelf in the background. The man is seated on the floor, with his arm resting on his knee. He is wearing a white sweater and blue pants. The woman is seated with her legs crossed, wearing a white blouse and blue overalls.}
\item{\textbf{Birthday:} A woman and a woman standing side by side, smiling at the camera. One wearing a white dress and the other in a pink dress. The woman on the left is holding a small cake. Behind them is a festive backdrop with a pink and white color scheme, adorned with streamers and bunting.}
\end{itemize}

\section{Illustrative Examples of the Dataset}
\label{appendix: Illustrative Examples of the Dataset}

The PairHuman dataset is organized into four topics: couples, wedding, female friends, and parent-child. It includes full-body, half-body, and close-up images that capture a variety of human actions and scenes. The dataset features diverse attire, such as formal wear, casual clothing, wedding dresses, and traditional Chinese garments such as Hanfu and qipao. Each image is accompanied by detailed captions that describe the individuals, their clothing, their actions, and the surrounding context to ensure comprehensive and accurate representation. The samples and descriptions for each category are provided in Tables E.1-E.4.

\section{Supplementary Dataset Statistics}
\label{appendix:supplementary_stats}

Table E.5 presents the frequency distribution of the top 30 clothing-related descriptors, offering a complementary view to Figure 5(f).

\begin{table*}[t]
\centering
\scriptsize
\caption{Sample images and descriptions for the Couples topic }
\renewcommand{\arraystretch}{1.8}
\begin{tabularx}{\textwidth}{ 
>{\centering}m{0.12\textwidth}  
m{0.33\textwidth} | 
>{\centering}m{0.12\textwidth} 
m{0.33\textwidth} } 
\hline
\textbf{Image} & \textbf{Description} & \textbf{Image} & \textbf{Description} \\
\hline
\vspace*{\fill}{\includegraphics[width=0.1\textwidth]{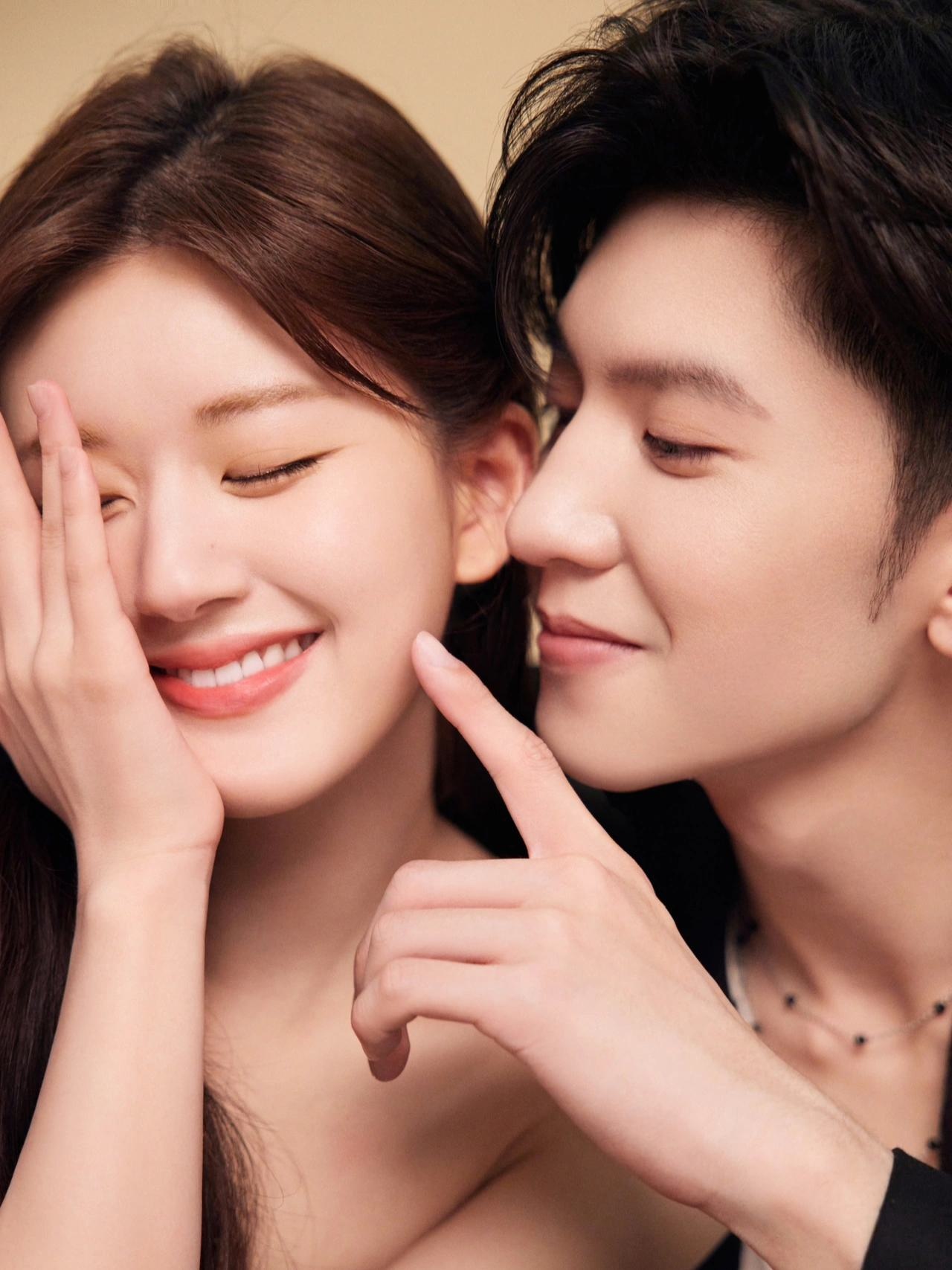}} & A close-up photograph of a man and a woman in a warm, intimate setting. The woman is smiling gently and has her hand near her face, while the man is smiling and has his hand near her ear. They are both looking at each other with affectionate expressions. The woman has long, dark hair and is wearing a light-colored top, while the man has short, dark hair and is wearing a dark jacket. The background is soft and blurred, emphasizing the subjects in the foreground. The overall mood of the image is one of happiness and closeness. &
\vspace*{\fill}{\includegraphics[width=0.1\textwidth]{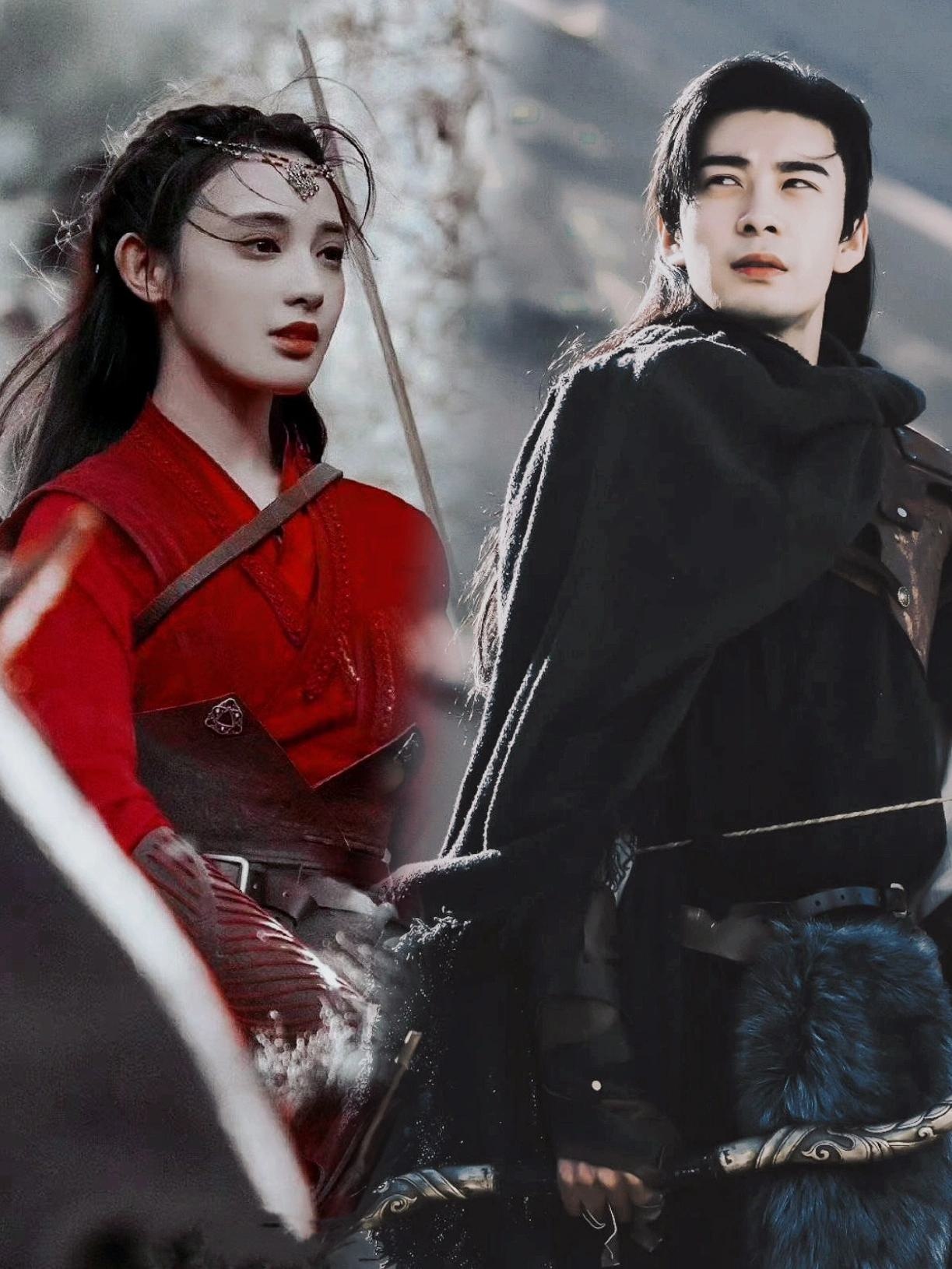}} & The image is a split-screen photograph featuring two individuals, likely from a historical or fantasy context, dressed in traditional Chinese attire. On the left, there is a woman with long dark hair, wearing a vibrant red outfit with a high collar and a headpiece that appears to be made of a light-colored material. She has a serious expression and her left hand is resting on her hip. On the right, there is a man with dark hair and a neutral expression. He is wearing a dark cloak with a fur-lined collar and a matching headpiece. The man's attire suggests a warrior or noble status, and he is holding a sword in his right hand, which is held at his side. The background of both images is a muted, overcast sky, which gives a dramatic effect to the overall composition. \\
\hline
\vspace*{\fill}{\includegraphics[width=0.1\textwidth]{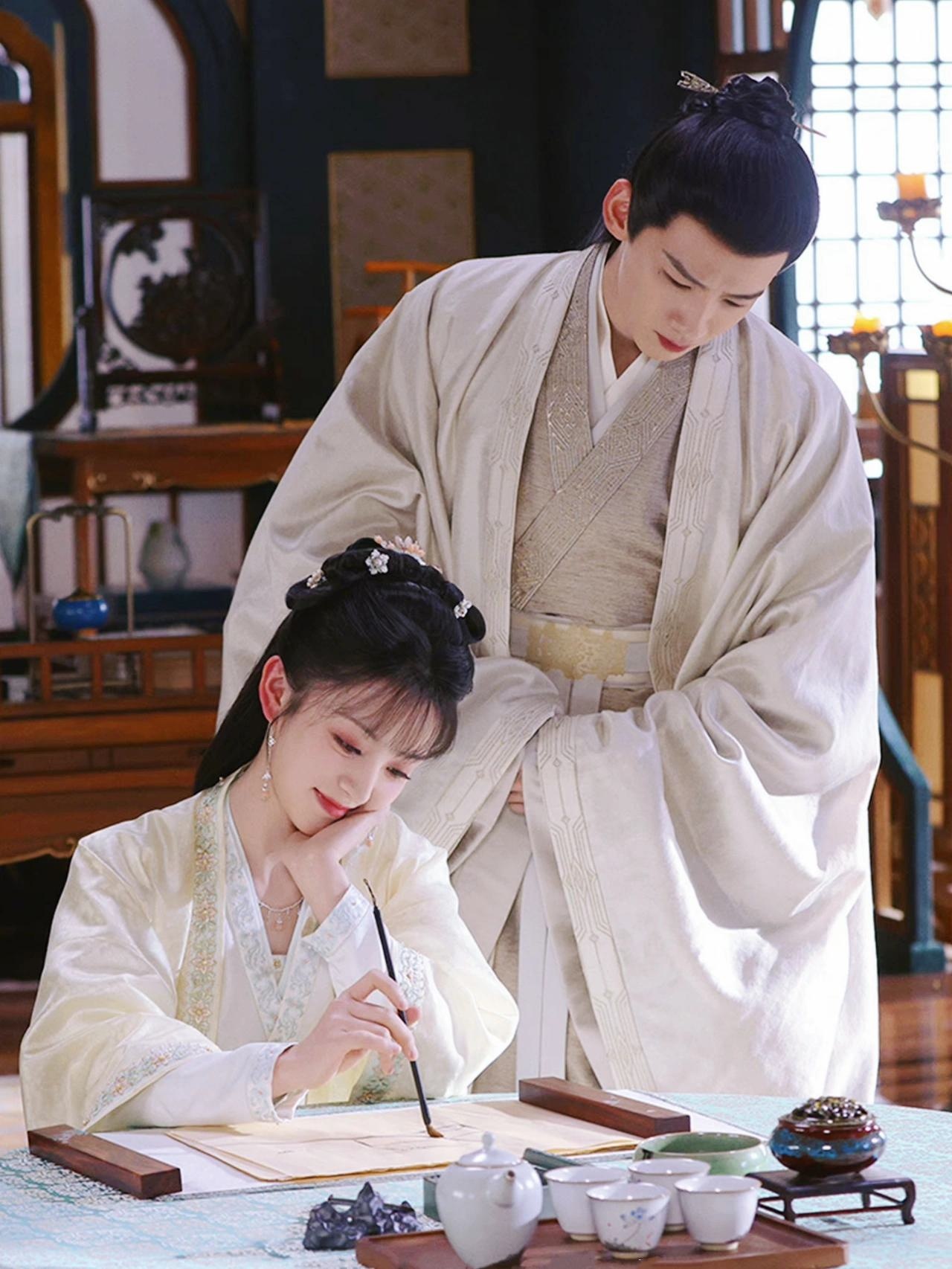}} & In this image, we see a scene from a traditional Chinese setting. A man and a woman, both dressed in traditional Chinese attire, are seated at a table. The man, on the right, is wearing a white robe with a high collar and a black belt, while the woman on the left is wearing a white robe with a floral pattern on the sleeves and a black belt. They are both engaged in an activity that appears to be writing or painting, as evidenced by the brush in the woman's hand and the inkwell on the table. The table is set with various items, including a teapot, a bowl, and a cup, suggesting a setting for a meal or a tea ceremony. The background features a wooden screen with intricate designs, adding to the traditional atmosphere of the scene. The overall mood of the image is serene and contemplative. &
\vspace*{\fill}{\includegraphics[width=0.1\textwidth]{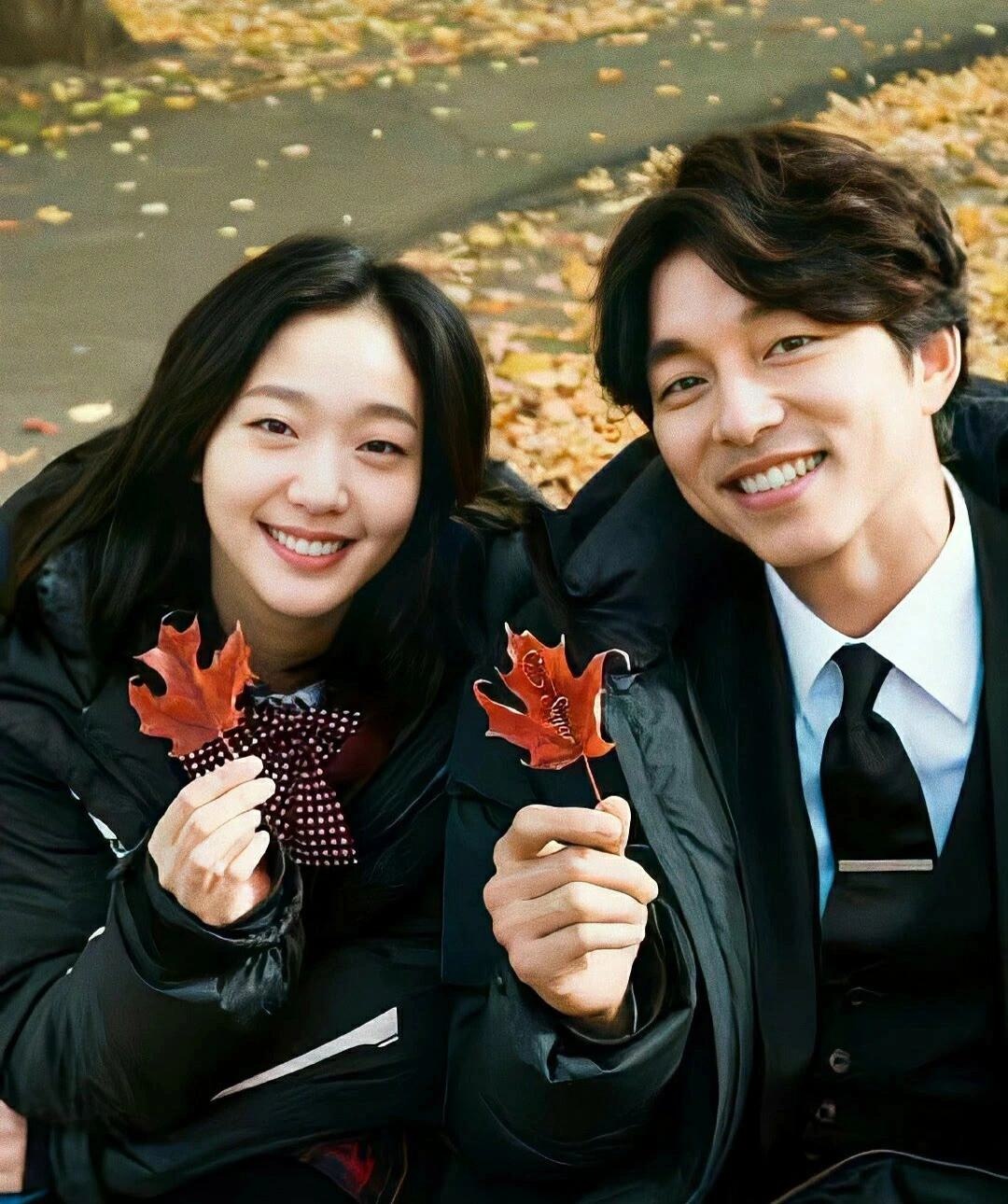}} & A young couple posing for a photo in a park during autumn. The woman is holding a red maple leaf and wearing a small red bow, while the man is holding a red maple leaf. They are both smiling and appear to be enjoying the moment. The couple is dressed in casual attire, with the woman wearing a black jacket and the man in a black suit with a white shirt and black tie. The background features a pathway lined with fallen leaves, suggesting a serene and picturesque setting. The overall mood of the image is warm and cheerful, capturing a moment of joy and togetherness.
 \\
\hline
\end{tabularx}
\end{table*}

\begin{table*}[t]
\centering
\scriptsize
\caption{Sample images and descriptions for the Wedding topic}
\renewcommand{\arraystretch}{1.8}
\begin{tabularx}{\textwidth}{ 
>{\centering}m{0.12\textwidth}  
m{0.33\textwidth} | 
>{\centering}m{0.12\textwidth} 
m{0.33\textwidth} } 
\hline
\textbf{Image} & \textbf{Description} & \textbf{Image} & \textbf{Description} \\
\hline
\vspace*{\fill}{\includegraphics[width=0.1\textwidth]{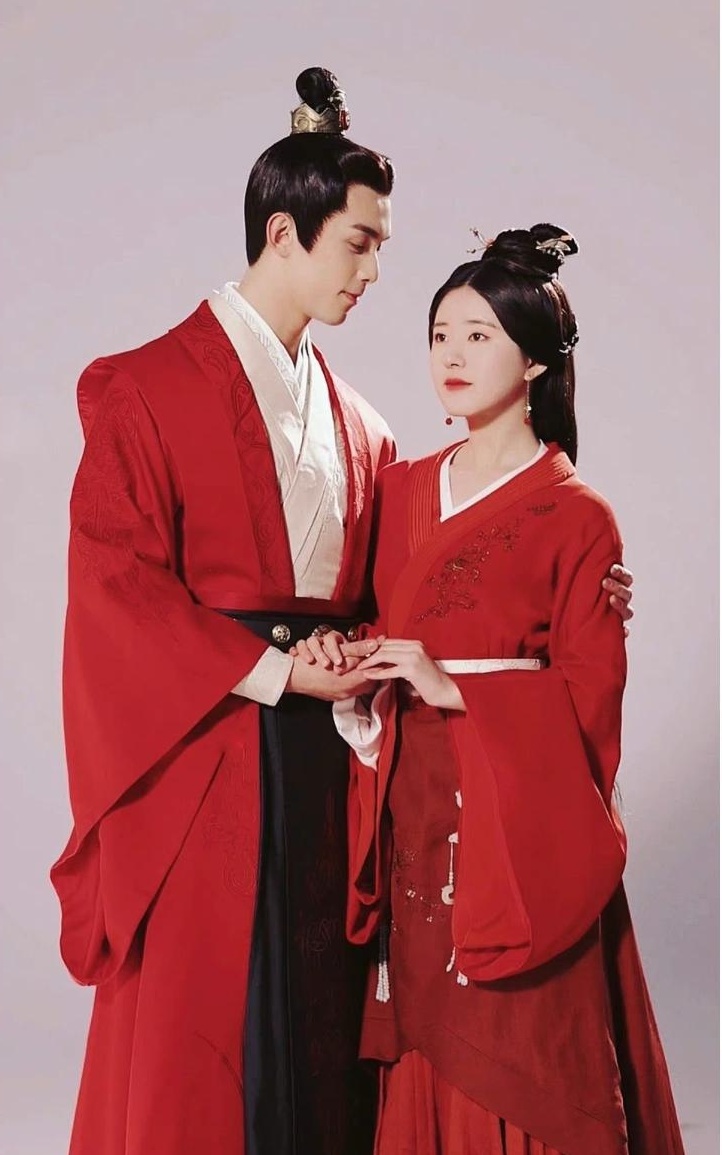}} & The image depicts a man and a woman dressed in traditional Chinese attire, likely from the Qing Dynasty period. The man is wearing a red robe with a high collar and a black belt, while the woman is dressed in a red dress with a white collar and a black belt. They are standing close to each other, with the woman's hand resting on the man's chest. Both individuals have their hair styled in a manner consistent with historical Chinese fashion. The background is a plain, light color. The lighting is soft and even. 
& \vspace*{\fill}{\includegraphics[width=0.1\textwidth]{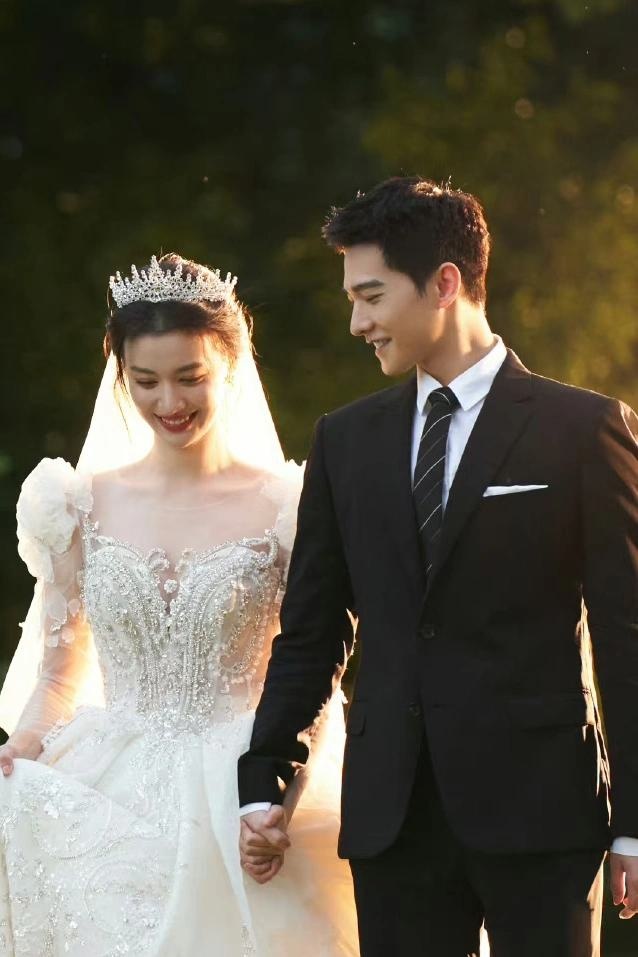}} & A bride and groom in traditional Chinese wedding attire, standing together outdoors during the day. The bride is wearing a white bridal gown with intricate beading and a veil, while the groom is dressed in a black suit with a white shirt and black tie. They are holding hands and smiling at the camera. The background is blurred, but it appears to be a natural outdoor setting with greenery and sunlight. The couple is the central focus of the image, and their attire and the act of holding hands suggest a moment of joy and celebration.
\\
\hline
\vspace*{\fill}{\includegraphics[width=0.1\textwidth]{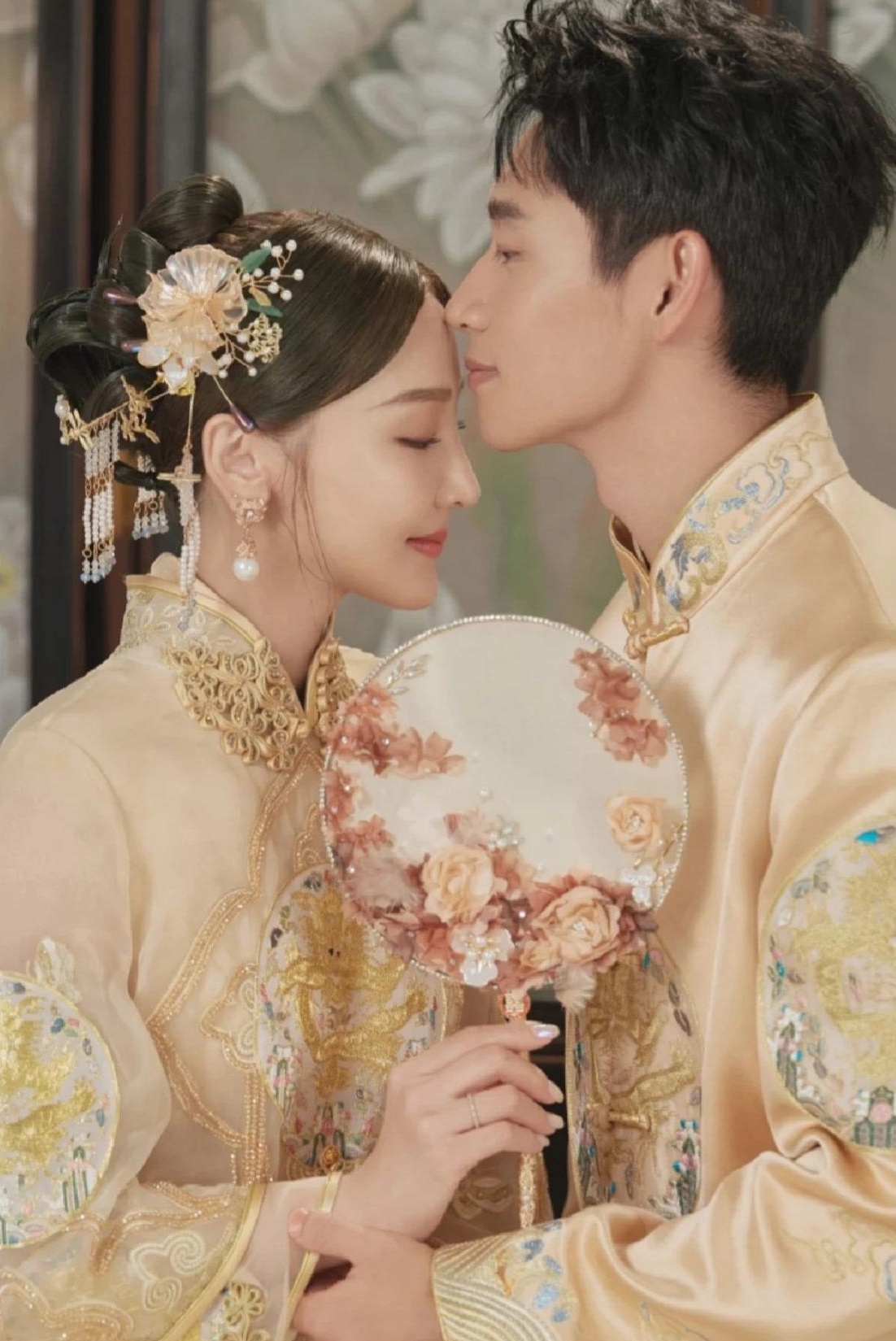}} & A young couple dressed in traditional Chinese attire, embracing and sharing a tender kiss. The man is wearing a gold-colored, ornate Chinese outfit with a high collar and intricate patterns, while the woman is adorned with a large, ornate hair accessory and a matching outfit. They are standing close together, with the man's hand gently resting on the woman's shoulder. The background features a soft, blurred floral pattern, suggesting a romantic or celebratory setting. The overall mood of the image is one of love and affection, with a focus on the couple's connection and the elegance of their traditional attire.
& \vspace*{\fill}{\includegraphics[width=0.1\textwidth]{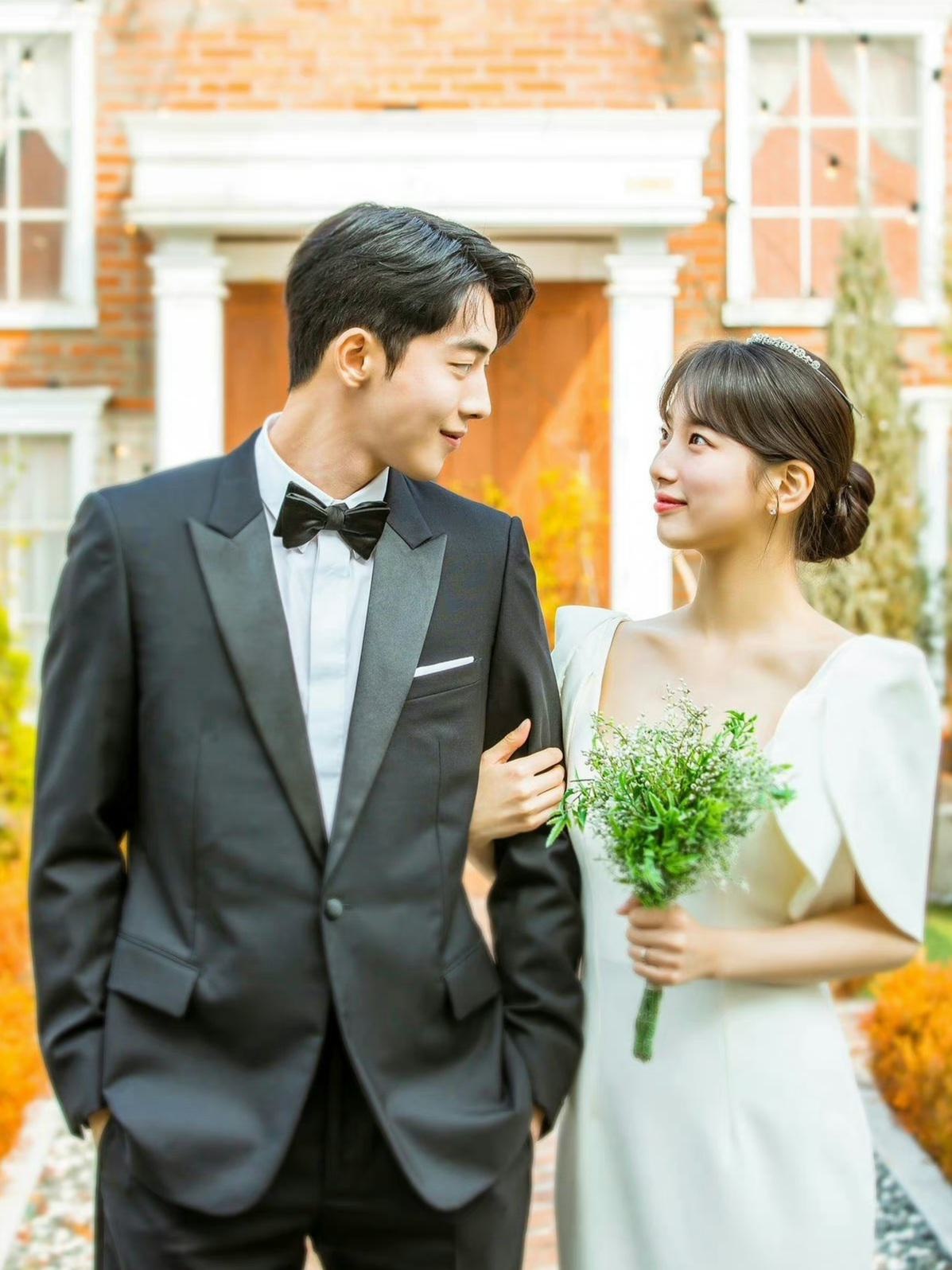}} & A young couple, dressed in formal attire, stand in front of a building with a red brick facade and white columns. The man is wearing a black tuxedo with a bow tie and white shirt, while the woman is in a white bridal gown with a floral bouquet in her hand. They are smiling and looking at each other, with the man's hand gently resting on the woman's arm. The setting is outdoors, likely during a wedding ceremony or a special event, given the couple's attire and the festive atmosphere. The background features a well-manicured garden with orange flowers, adding a touch of color to the scene. The overall mood of the image is joyful and celebratory, capturing a moment of happiness and love.
 \\
\hline
\end{tabularx}
\end{table*}

\begin{table*}[t]
\centering
\scriptsize
\caption{Sample images and descriptions for the Female Friends topic}
\renewcommand{\arraystretch}{1.8}
\begin{tabularx}{\textwidth}{ 
>{\centering}m{0.12\textwidth}  
m{0.33\textwidth} | 
>{\centering}m{0.12\textwidth} 
m{0.33\textwidth} } 
\hline
\textbf{Image} & \textbf{Description} & \textbf{Image} & \textbf{Description} \\
\hline
\vspace*{\fill}{\includegraphics[width=0.1\textwidth]{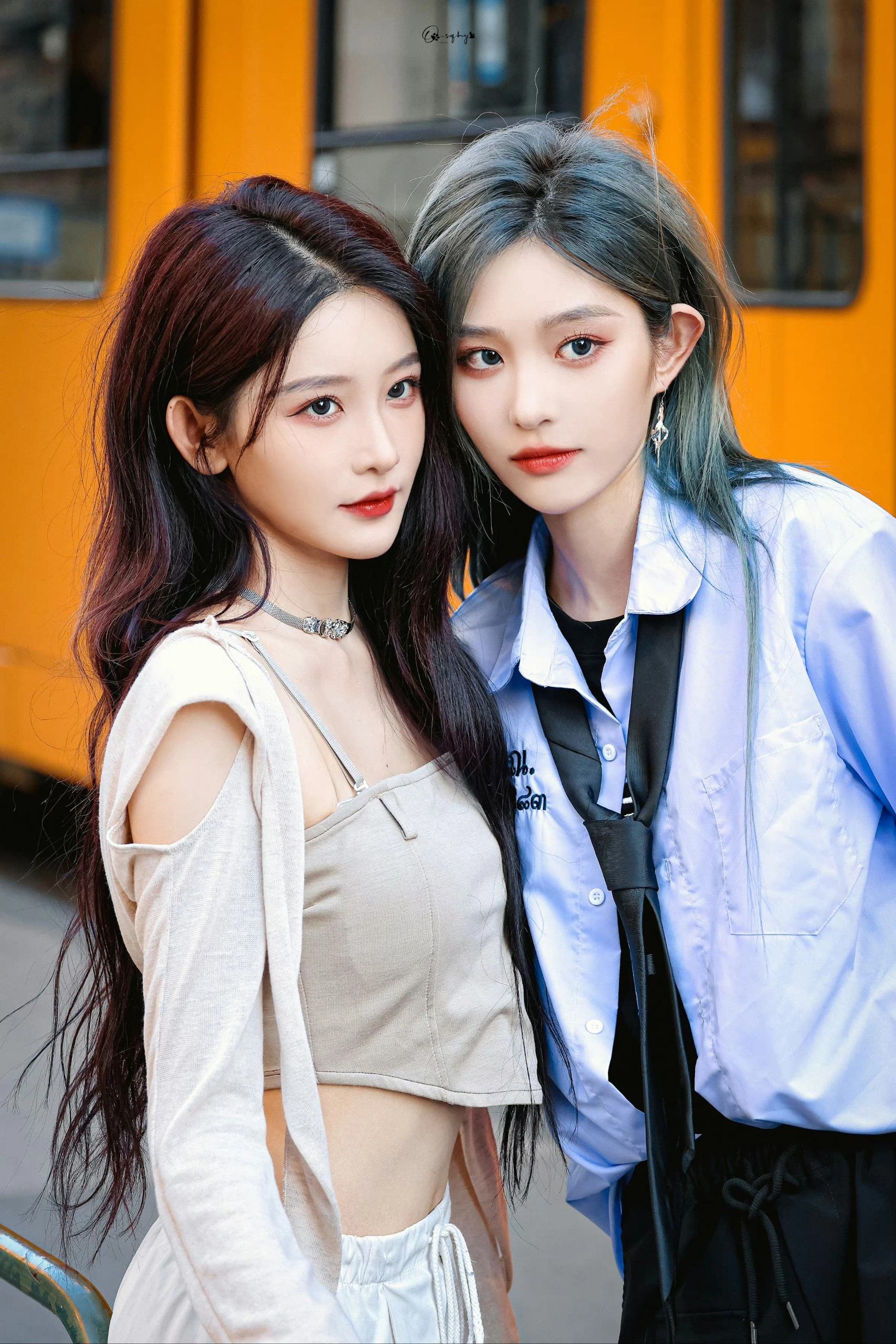}} & Two young women posing for a photo. The woman on the left has long, dark hair and is wearing a beige off-the-shoulder top with a high neckline. She has a neutral expression and is looking directly at the camera. The woman on the right has short, light blue hair and is wearing a white shirt with a black tie. She has a subtle smile and is also looking at the camera. They are standing close to each other, with the woman on the right slightly leaning towards the woman on the left. In the background, there is a bright orange structure that appears to be a part of a public transportation vehicle, possibly a bus or train. The setting suggests an urban environment, and the mood of the image is casual and friendly.
& \vspace*{\fill}{\includegraphics[width=0.1\textwidth]{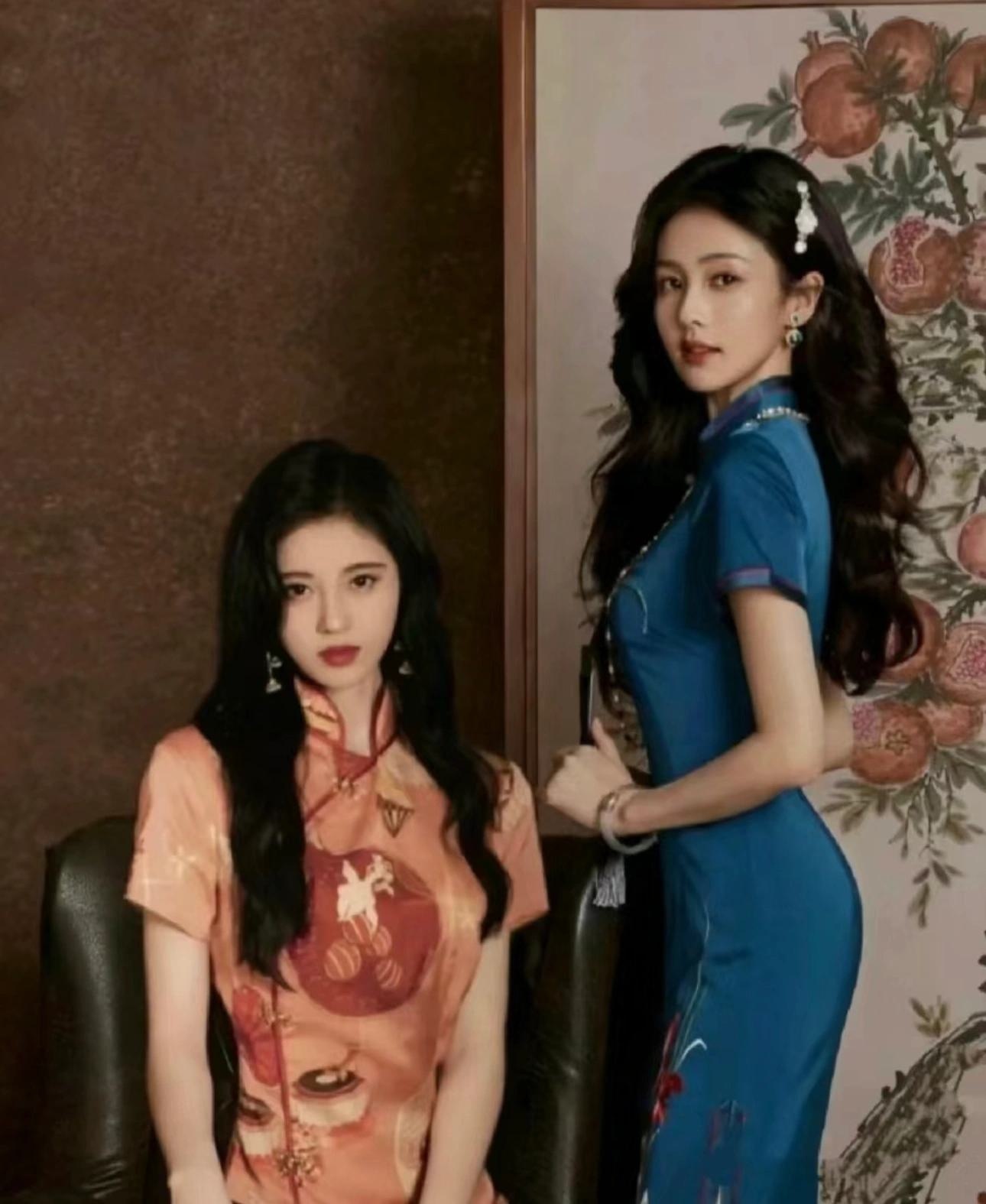}} & Two women in traditional Chinese attire, posing for a portrait. The woman on the left is seated, wearing a vibrant orange dress with a floral pattern and a red hair accessory. She has long black hair and is looking directly at the camera with a neutral expression. The woman on the right is standing, wearing a dark blue dress with a floral pattern and a white hair accessory. She has long black hair and is also looking directly at the camera with a neutral expression. They are positioned in front of a wall with a floral pattern and a wooden frame. The overall mood of the image is serene and elegant, capturing a moment of cultural heritage and beauty.
 \\
\hline
\vspace*{\fill}{\includegraphics[width=0.1\textwidth]{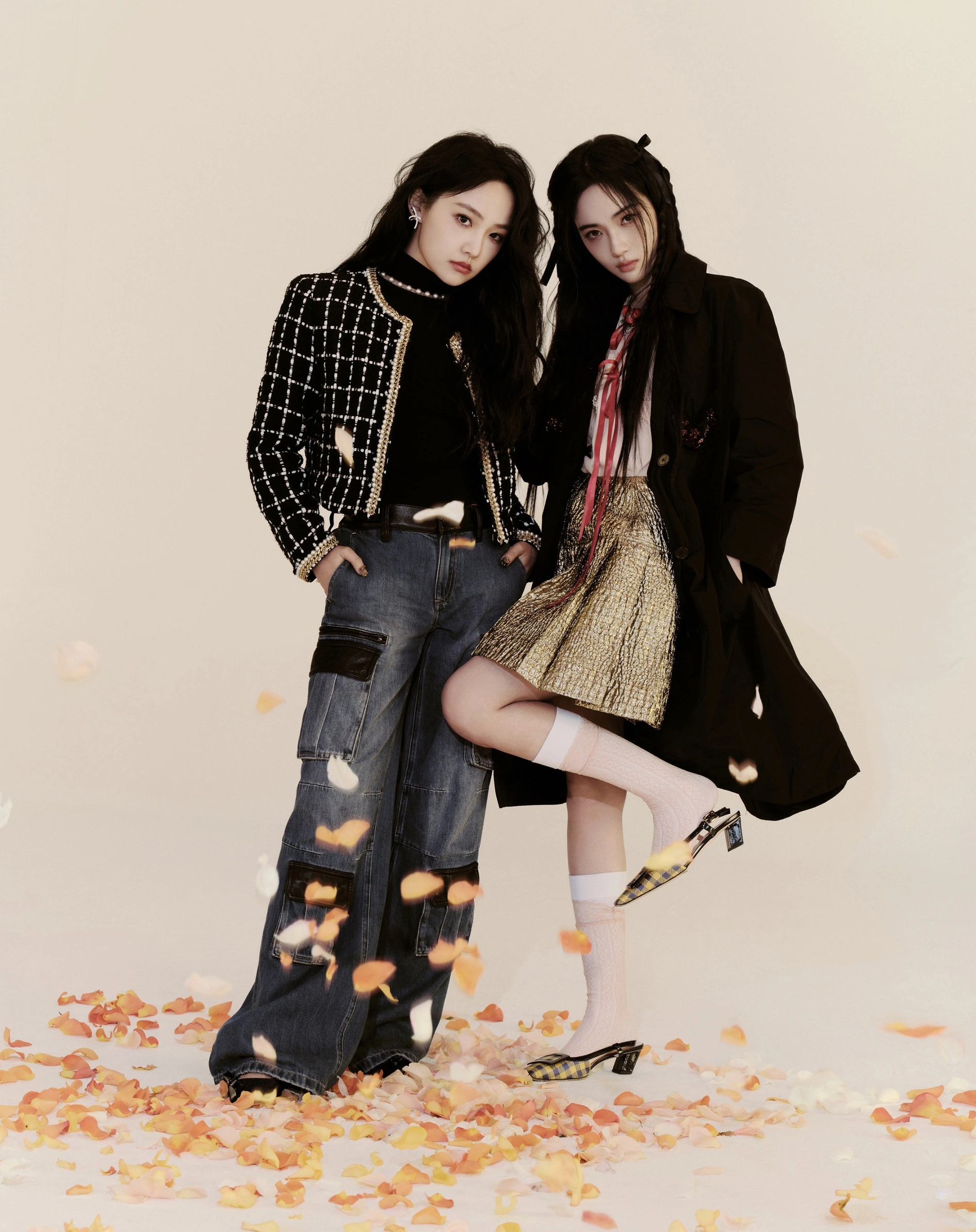}} & Two women pose for a fashion-forward photograph. The woman on the left is dressed in a black and white checkered jacket over a black turtleneck, paired with high-waisted blue jeans. She has her left hand on her hip and her right hand resting on her thigh. The woman on the right is wearing a black coat over a white blouse with a red scarf, a gold skirt, and black high heels. She has her right hand on her hip and her left hand resting on her thigh. Both women are looking directly at the camera with a neutral expression. They are standing on a bed of orange petals, which are scattered around them. The background is a plain, light beige color, providing a neutral backdrop that highlights the subjects. The overall mood of the image is elegant and fashionable, with a focus on the clothing and the interaction between the two women.
& \vspace*{\fill}{\includegraphics[width=0.1\textwidth]{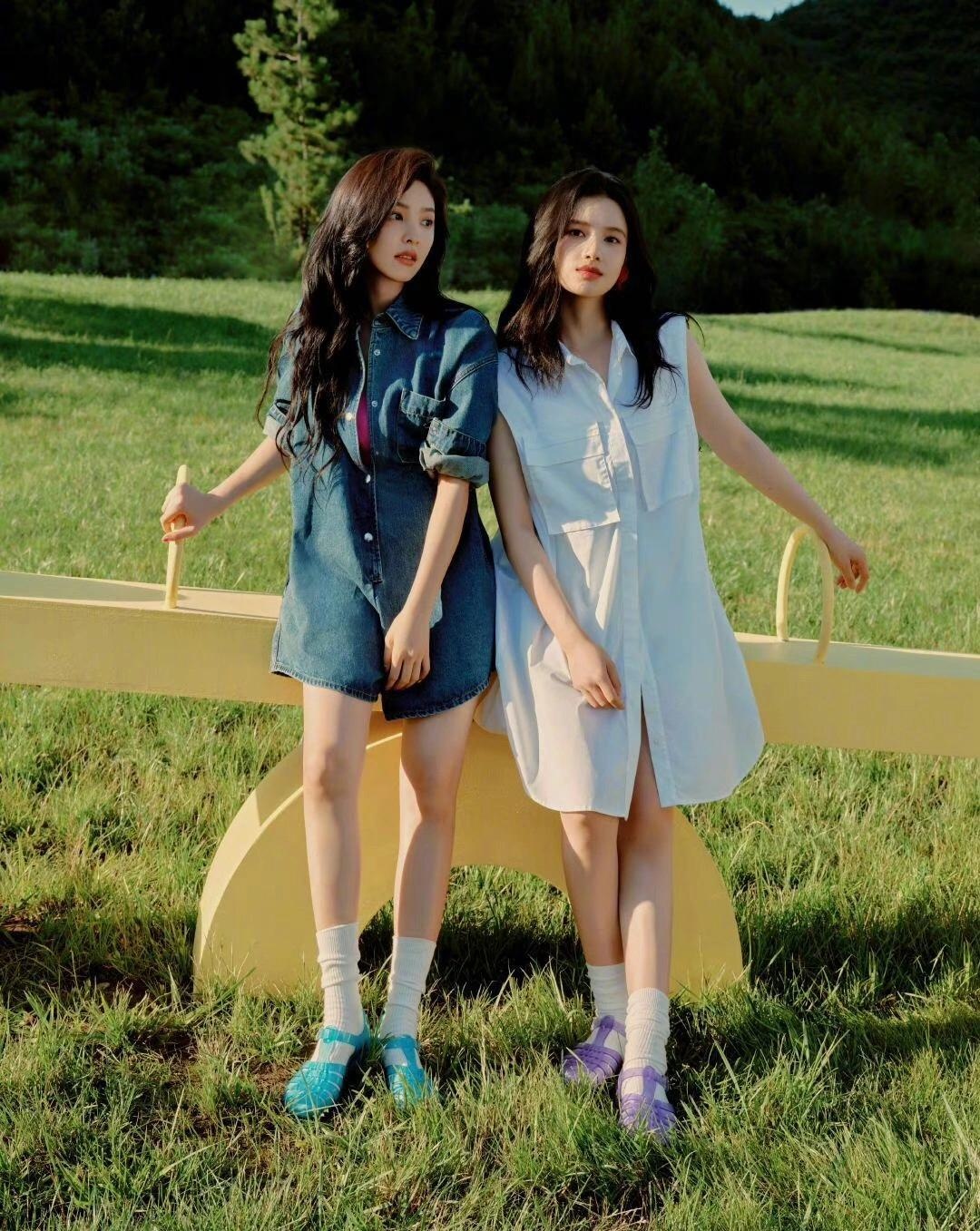}} & Two women posing for a photo in a grassy field. The woman on the left is wearing a denim dress and blue shoes, while the woman on the right is dressed in a white blouse and purple shoes. Both women are standing on a yellow bench, which is positioned in the center of the image. The background features a lush green field and a line of trees. The overall mood of the image is relaxed and casual, with a focus on the contrast between the denim and white clothing styles. The photo style appears to be a casual, outdoor portrait with a shallow depth of field, emphasizing the subjects in the foreground.
 \\
\hline
\end{tabularx}
\end{table*}

\begin{table*}[t]
\centering
\scriptsize
\caption{Sample images and descriptions for the Parent-Child topic}
\renewcommand{\arraystretch}{1.8}
\begin{tabularx}{\textwidth}{ 
>{\centering}m{0.12\textwidth}  
m{0.33\textwidth} | 
>{\centering}m{0.12\textwidth} 
m{0.33\textwidth} } 
\hline
\textbf{Image} & \textbf{Description} & \textbf{Image} & \textbf{Description} \\
\hline
\vspace*{\fill}{\includegraphics[width=0.1\textwidth]{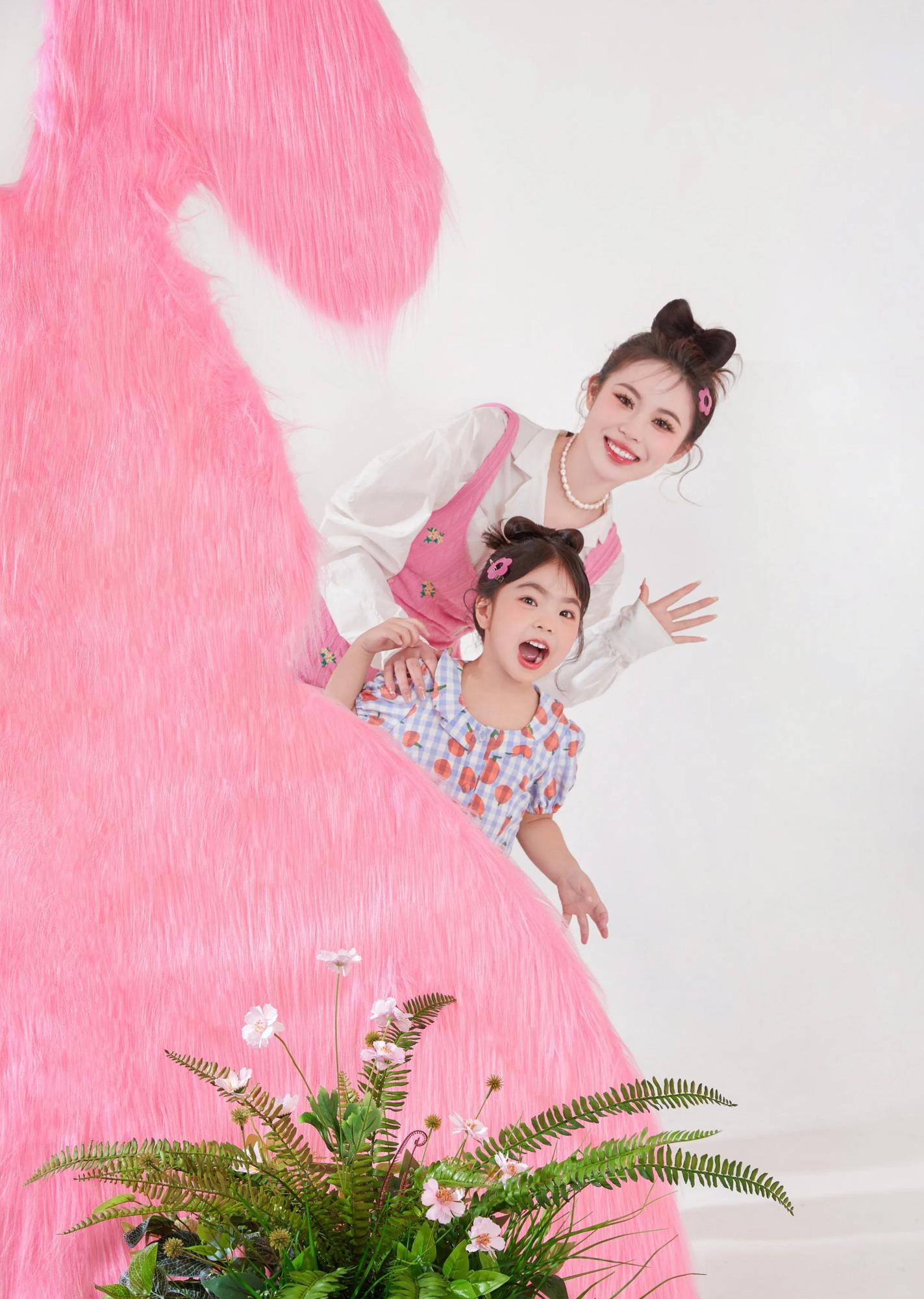}} & The image features a joyful scene with two individuals, likely a mother and daughter, standing in front of a large, fluffy pink prop. The mother is on the left, holding the child's hand, and both are smiling broadly. The child is wearing a light blue dress with a floral pattern and has a playful expression. They are positioned in front of a white wall, which contrasts with the vibrant pink of the prop. To the right of the frame, there is a potted plant with green leaves and pink flowers, adding a touch of nature to the composition. The overall mood of the image is cheerful and lighthearted, with a playful and whimsical theme.
& \vspace*{\fill}{\includegraphics[width=0.1\textwidth]{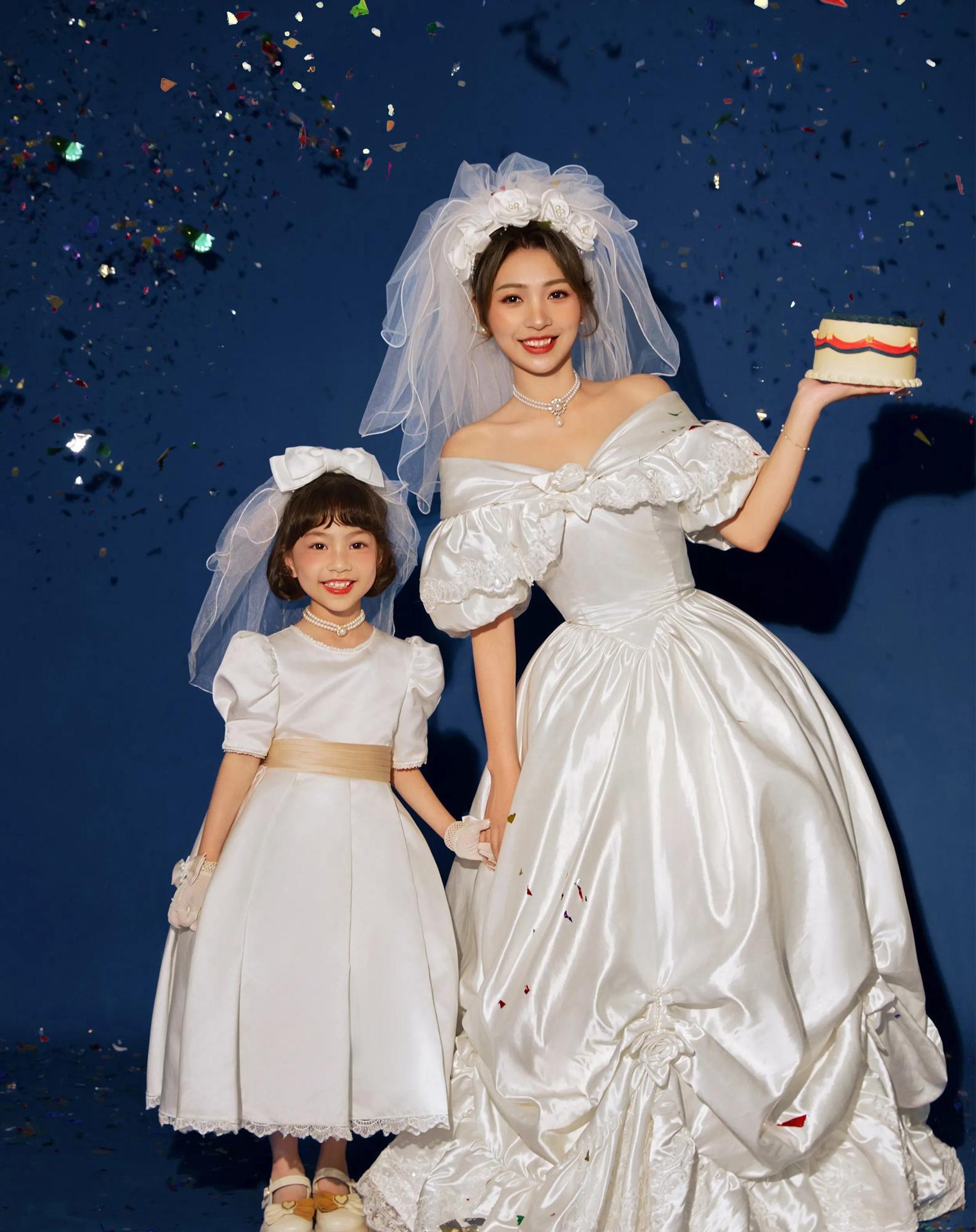}} & In the image, a bride and a young girl are the main subjects. The bride is elegantly dressed in a white wedding gown with a long train, and she is holding a small cake with a red and white striped ribbon. The young girl, wearing a matching white dress with a bow in her hair, is holding the bride's hand. They are standing against a blue background with confetti scattered around them, suggesting a celebratory occasion. The bride is smiling and appears to be in a joyful mood, while the young girl is looking at the camera with a slight smile. The photo style is a posed portrait with a focus on the subjects, and the clothing style is formal and traditional.
 \\
\hline
\vspace*{\fill}{\includegraphics[width=0.1\textwidth]{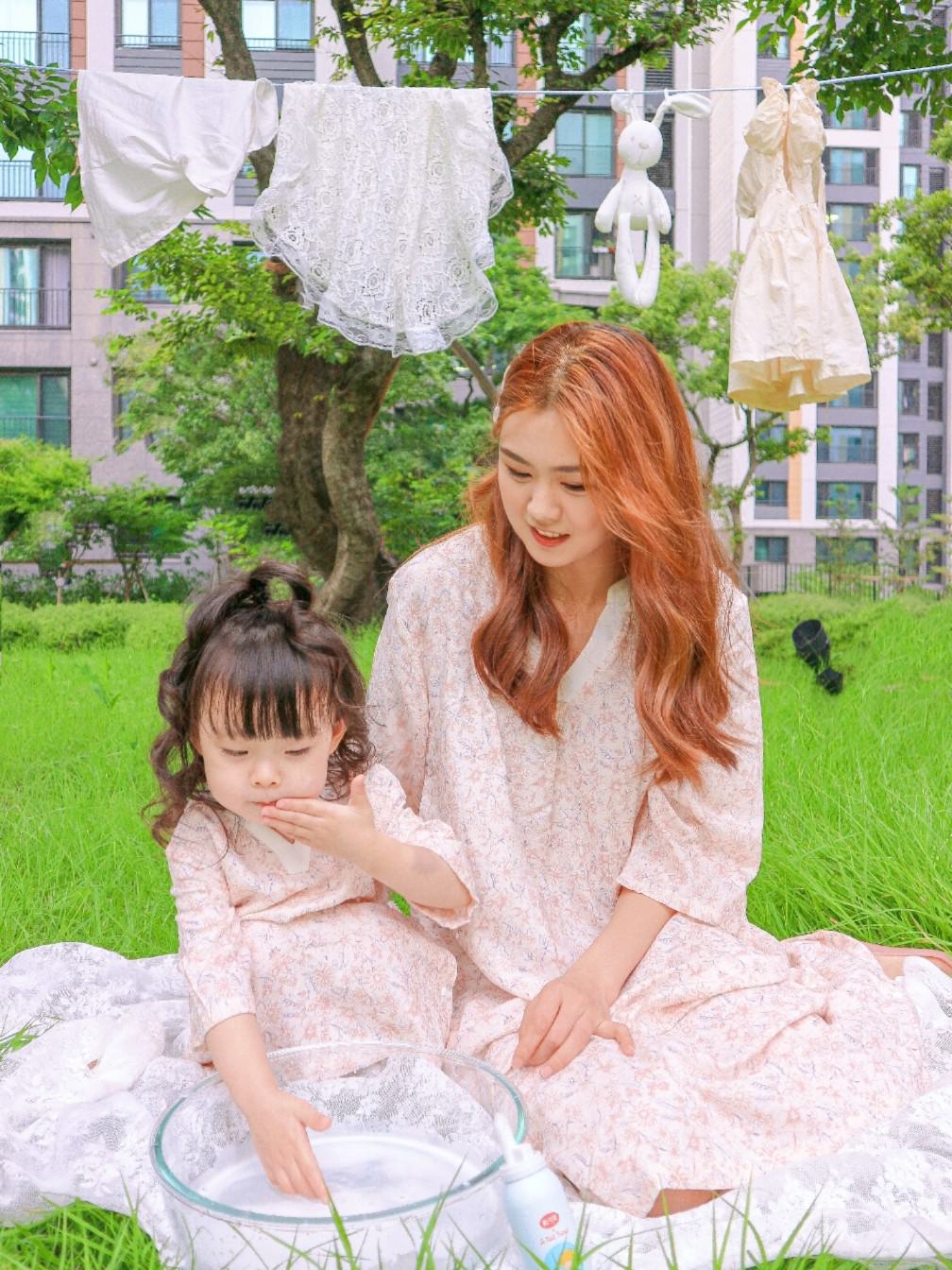}} & A young girl and an adult woman are seated on the grass under a tree. The girl is wearing a light pink dress and is holding a small white bowl. The woman is wearing a matching light pink dress and is looking at the girl with a smile. They are surrounded by laundry hanging on a clothesline, and there are white bunnies hanging from the line. The setting appears to be a park or a garden, and the mood of the image is serene and joyful.
& \vspace*{\fill}{\includegraphics[width=0.1\textwidth]{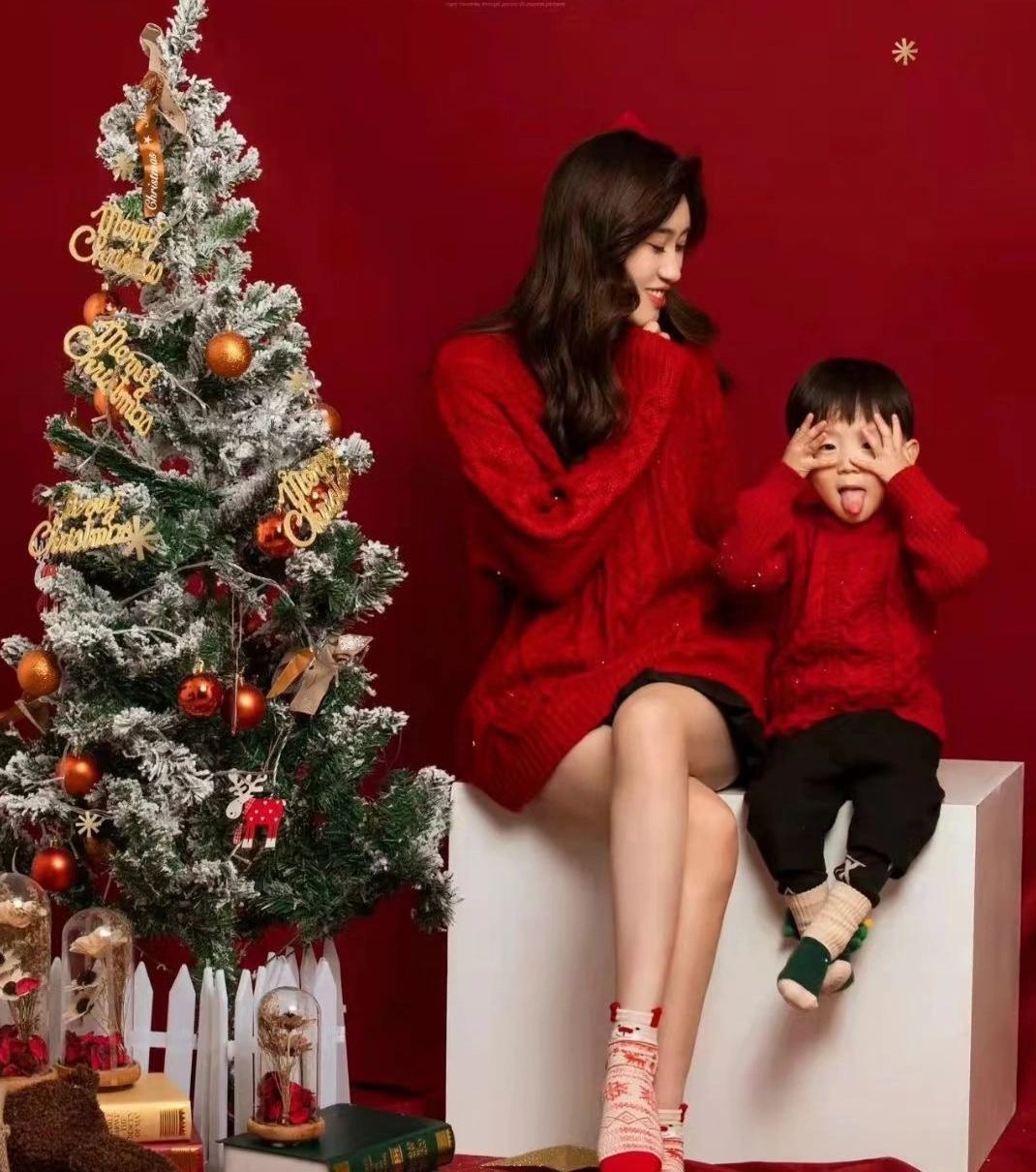}} & A festive holiday scene with a woman and a child. The woman, wearing a red sweater and sitting on a white bench, is smiling and looking at the child. The child, dressed in a red sweater and black pants, is playfully covering his eyes with his hands. They are in front of a beautifully decorated Christmas tree adorned with ornaments and lights. The tree is placed on a red carpet, and there are presents and a teddy bear nearby. The setting appears to be indoors, with a joyful and lighthearted atmosphere, likely during the holiday season. 
 \\
\hline
\end{tabularx}
\end{table*}

\begin{table}[htbp]
  \centering
  \caption{Frequency distribution of the top 30 clothing-related descriptors, including adjectives grouped by category, clothing nouns, and adjective–noun combinations.}
  \label{tab:categorized_adj_noun_matching}
  \renewcommand{\arraystretch}{1.1} 
  \adjustbox{max width=\linewidth}{%
    \begin{tabular}{llllllll} 
      \toprule
      \multicolumn{3}{c}{\textbf{Categorized Adjectives}} & \multicolumn{2}{c}{\textbf{Nouns}} & \multicolumn{3}{c}{\textbf{Adjective-Noun Pairs}} \\
      \cmidrule(lr){1-3} \cmidrule(lr){4-5} \cmidrule(lr){6-8}
      Category & Adjective & Count & Noun & Count & Matching Pair & Count \\
      \midrule
      \multirow{16}{*}{Color} & white & 69803 & dress & 40902 & white dress & 18156 \\
      & black & 44257 & suit & 35832 & black suit & 17307 \\
      & red & 15098 & shirt & 16312 & white shirt & 13721 \\
      & dark & 4892 & sweater & 7339 & white suit & 9854 \\
      & blue & 3228 & bridal & 6910 & black dress & 7161 \\
      & gray & 2947 & top & 6350 & white bridal & 6815 \\
      & green & 2443 & bow & 5788 & red dress & 6174 \\
      & light & 1783 & tuxedo & 5294 & black tuxedo & 4493 \\
      & brown & 1574 & blouse & 4711 & white blouse & 4296 \\
      & yellow & 1380 & outfit & 4321 & black bow & 3891 \\
      & beige & 752 & blazer & 3635 & dark suit & 3511 \\
      & purple & 211 & robe & 3595 & traditional dress & 3436 \\
      & colorful & 212 & t-shirt & 3498 & white sweater & 3128 \\
      & gold & 159 & tie & 3230 & white top & 2944 \\
      & pink & 113 & turtleneck & 1609 & white t-shirt & 2879 \\
      & navy & 79 & tank & 1070 & red sweater & 2724 \\
      
      \addlinespace[-\belowrulesep]
      \cmidrule(lr){1-3}\addlinespace[-\belowrulesep]
      
      \multirow{4}{*}{Theme} & traditional & 5607 & coat & 1059 & gray suit & 2245 \\
      & formal & 734 & hat & 992 & red robe & 2153 \\
      & chinese & 441 & garment & 944 & black tie & 1887 \\
      & intricate & 34 & skirt & 866 & black top & 1732 \\

      \addlinespace[-\belowrulesep]
      \cmidrule(lr){1-3}\addlinespace[-\belowrulesep]
      
      \multirow{6}{*}{Style} & floral & 1619  & vest & 803 & traditional outfit & 1491 \\
      & ruffled & 437 & gown & 769 & black blazer & 1350 \\
      & casual & 206 & pant & 527 & white blazer & 1299 \\
      & sailor-style & 113 & veil & 469 & floral dress & 1173 \\
      & striped & 89 & uniform & 454 & black turtleneck & 1126 \\
      & checkered & 31 & shoe & 332 & black shirt & 1108 \\
      
      \addlinespace[-\belowrulesep]
      \cmidrule(lr){1-3}\addlinespace[-\belowrulesep]
      
      \multirow{4}{*}{Material} & beaded & 93  & belt & 309 & green dress & 1095 \\
      & fluffy & 53 & boutonniere & 183 & red outfit & 996 \\
      & feathered & 49 & emblem & 171 & blue dress & 924 \\
      & fur & 16 & bodice & 125 & white bow & 815 \\
      \bottomrule
    \end{tabular}
  }
\end{table}

\bibliographystyle{elsarticle-num}
\bibliography{pairhuman_new}

\end{document}